\documentclass[10pt,twocolumn,letterpaper]{article}

\ifdefined\pdfminorversion\pdfminorversion=7\fi

\newif\ifarxiv
\arxivtrue

\ifarxiv
\usepackage[pagenumbers]{cvpr}
\usepackage{fontawesome5}
\else
\usepackage[review]{cvpr}
\fi

\usepackage{amsthm}
\usepackage{needspace}
\usepackage{bm}
\usepackage{multirow}
\usepackage{colortbl}
\newcolumntype{L}[1]{>{\raggedright\arraybackslash}p{#1}}
\usepackage{algorithm}
\usepackage{algorithmic}

\definecolor{oursrow}{RGB}{240,246,254}
\definecolor{commentcolor}{RGB}{115,115,115}

\newcommand{\dashedmidrule}[1]{%
  \noalign{\vskip1.6pt}%
  \multispan{#1}\leaders\hbox to 4pt{\hss\vrule height.35pt width2pt\hss}\hfill\\%
  \noalign{\vskip1.6pt}%
}

\newcommand{\ours}{\texttt{\mdseries eDMD}\xspace}
\newcommand{\sdthree}{SD3-Medium\xspace}
\newcommand{\flux}{FLUX.2-klein-4B\xspace}
\newcommand{\fluxnine}{FLUX.2-klein-9B\xspace}
\newcommand{\qwen}{Qwen-Image\xspace}
\newcommand{\hpspp}{HPSv3++\xspace}

\DeclareMathOperator{\KL}{KL}
\newcommand{\E}{\mathbb{E}}
\newcommand{\sg}{\operatorname{sg}}
\newcommand{\x}{\mathbf{x}}
\newcommand{\y}{\mathbf{y}}
\newcommand{\z}{\mathbf{z}}
\newcommand{\Y}{\mathbf{Y}}
\newcommand{\xh}{\hat{\mathbf{x}}}
\newcommand{\eps}{\bm{\epsilon}}
\newcommand{\Tc}{T_{\mathrm{c}}}
\newcommand{\Tu}{T_{\mathrm{u}}}
\newcommand{\Tw}{T^{w}}

\theoremstyle{plain}
\newtheorem{proposition}{Proposition}
\newtheorem{lemma}{Lemma}
\theoremstyle{definition}

\theoremstyle{remark}

\AtEndPreamble{%
  \crefname{algorithm}{Alg.}{Algs.}%
  \Crefname{algorithm}{Algorithm}{Algorithms}%
  \crefname{proposition}{Prop.}{Props.}%
  \crefname{lemma}{Lemma}{Lemmas}%
  \Crefname{lemma}{Lemma}{Lemmas}%
  \Crefname{proposition}{Proposition}{Propositions}%
  \crefname{definition}{Def.}{Defs.}%
  \crefname{figure}{Fig.}{Figs.}%
  \crefname{equation}{Eq.}{Eqs.}%
  \crefname{appendix}{App.}{Apps.}%
}

\definecolor{cvprblue}{rgb}{0.21,0.49,0.74}
\usepackage[pagebackref,breaklinks,colorlinks,hypertexnames=false,allcolors=cvprblue]{hyperref}

\def\confName{CVPR}
\def\confYear{2027}

\title{\ours: Endpoint Distribution Matching for Few-Step Diffusion Distillation}

\ifarxiv
\author{%
Tong Zhao$^{1,2,3}$ \quad
Xianfang Zeng$^{3}$ \quad
Yucheng Han$^{3}$ \quad
Ruoyu Wang$^{2}$\\
Mingkun Lei$^{3}$ \quad
Gang Yu$^{3}$ \quad
Chi Zhang$^{2,\dagger}$\\[6pt]
{\normalsize $^{1}$Zhejiang University \quad $^{2}$AGI Lab, Westlake University \quad $^{3}$StepFun}\\[6pt]
{\normalsize\faGithub\hspace{0.5em}\href{https://github.com/Westlake-AGI-Lab/eDMD}{\texttt{https://github.com/Westlake-AGI-Lab/eDMD}}}
}
\else
\author{Anonymous Authors\\
Anonymous Institution\\
{\tt\small anonymous@example.com}
}
\fi

\makeatletter
\long\def\appendixcleanentry\numberline#1\hskip -1em.~#2\@nil{\numberline{#1}#2}
\long\def\appendixplainentry#1\@nil{#1}
\newcommand{\appendixformatentry}{\@ifnextchar\numberline{\appendixcleanentry}{\appendixplainentry}}
\newcommand{\appendixcontents}{%
  \let\appendixoriginalcontentsline\contentsline
  \def\contentsline##1##2##3##4{%
    \appendixoriginalcontentsline{##1}{\appendixformatentry##2\@nil}{##3}{##4}}%
  \renewcommand*{\l@section}[2]{%
    \addvspace{3pt}%
    \@dottedtocline{1}{0em}{1.5em}{\bfseries##1}{\bfseries##2}}%
  \renewcommand*{\l@subsection}[2]{\@dottedtocline{2}{1em}{2.5em}{##1}{##2}}%
  \@starttoc{toc}%
}
\makeatother

\begin{document}
\ifarxiv
\addtocontents{toc}{\protect\setcounter{tocdepth}{-1}}
\fi
\twocolumn[{%
\renewcommand\twocolumn[1][]{#1}%
\maketitle
\begin{center}
\vspace{-5mm}%
\captionsetup{type=figure}
\includegraphics[width=\textwidth]{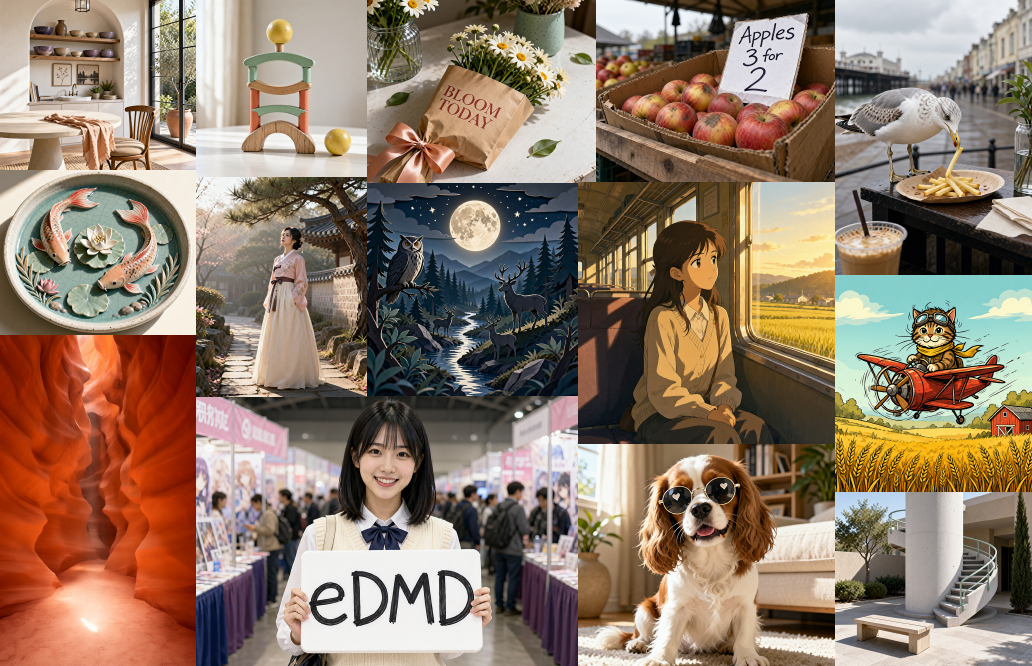}
\captionof{figure}{\textbf{4-step samples from} \ours, \textbf{distilled from \qwen-2.1.} Using a tenth of the teacher's 40 steps, \ours renders photorealistic scenes, legible text and diverse artistic styles.}%
\label{fig:teaser}
\end{center}%
\vspace{0.4em}
}]
\ifarxiv
{\renewcommand{\thefootnote}{}\footnotetext{$^{\dagger}$Corresponding author: \texttt{chizhang@westlake.edu.cn}.}}
\fi
\begin{abstract}
Distribution matching distillation (DMD) has emerged as a leading paradigm for distilling diffusion models into few-step students, where an auxiliary fake score model tracks the student's distribution and the discrepancy between the teacher and fake predictions drives the student toward the teacher.
In multi-step distillation, although only the student's final output is delivered, existing methods train the fake model on intermediate states or across all denoising steps.
Consequently, the student's update signal becomes entangled with intermediate distributions, which can distort the matching direction and remain nonzero even after the student's final outputs already match the teacher distribution.
To resolve this mismatch, we propose Endpoint Distribution Matching Distillation (\ours), which trains a single shared fake model exclusively on the student's final outputs and uses it to correct every denoising step.
Our theoretical analysis proves that this endpoint reference guarantees that the distribution matching term vanishes if and only if the student's final output distribution matches the teacher's, regardless of intermediate distributions.
Beyond the fake reference, we further strengthen teacher supervision through a precise teacher (PT) that replaces single-step guided predictions with short multi-step trajectories on the final noise interval, paired with a calibrated query density for intermediate steps.
Experiments on \sdthree, FLUX.2-klein (4B and 9B), and \qwen-2.1 demonstrate that \ours consistently outperforms existing distribution matching methods for 4-step generation.%
\end{abstract}

\section{Introduction}
\label{sec:intro}

Diffusion and flow matching models have achieved remarkable success in high-quality image~\cite{ho2020ddpm,esser2024sd3}, video~\cite{wan2025wan}, and audio~\cite{liu2023audioldm} generation.
However, their iterative sampling processes typically require tens of network evaluations, resulting in substantial inference costs.
Few-step distillation addresses this limitation by transferring the generative capabilities of a pretrained teacher into a student that requires only one or a few denoising steps~\cite{salimans2022progressive,song2023consistency,ren2024hypersd,lin2024sdxllightning}.
Among existing approaches, distribution matching distillation (DMD) has emerged as one of the most effective and widely adopted paradigms~\cite{luo2023diffinstruct,yin2024dmd}.
While DMD has demonstrated impressive one-step generation capabilities, a single network evaluation often struggles to preserve fine-grained details and complex structures~\cite{yin2024dmd2}.
Recent methods therefore extend DMD to multi-step generation, allowing the student to progressively refine its predictions and achieve higher generation quality while maintaining efficient inference~\cite{yin2024dmd2,luo2025tdm,liu2026ddmd,liu2026cdm}.\looseness=-1

The central idea of DMD is to train the student by minimizing the distributional discrepancy between its generated samples and those of the pretrained teacher.
Specifically, DMD formulates this objective through the reverse Kullback--Leibler (KL) divergence between their noise-perturbed distributions, whose gradient can be estimated from the difference between the teacher's and student's score functions~\cite{yin2024dmd}.
Since the student's score is not directly available, DMD introduces an auxiliary diffusion model, known as the \emph{fake score model}, which is trained online on student-generated samples to approximate the student's distribution.
The discrepancy between the teacher's prediction, often enhanced with classifier-free guidance (CFG)~\cite{ho2022cfg}, and the fake model's estimate then provides the correction signal for updating the student.
Importantly, the distribution learned by the fake model determines the reference against which the student is optimized.
When denoising is performed in a single step, the student's prediction is also its final generated output, making the distribution learned by the fake model naturally consistent with the desired matching objective.
When denoising involves multiple steps, however, the student produces a sequence of intermediate predictions before generating its final output.
These predictions generally follow different distributions, so how the fake model is trained becomes a central design choice.\looseness=-1

\begin{figure*}[t]
\centering
\includegraphics[width=\textwidth]{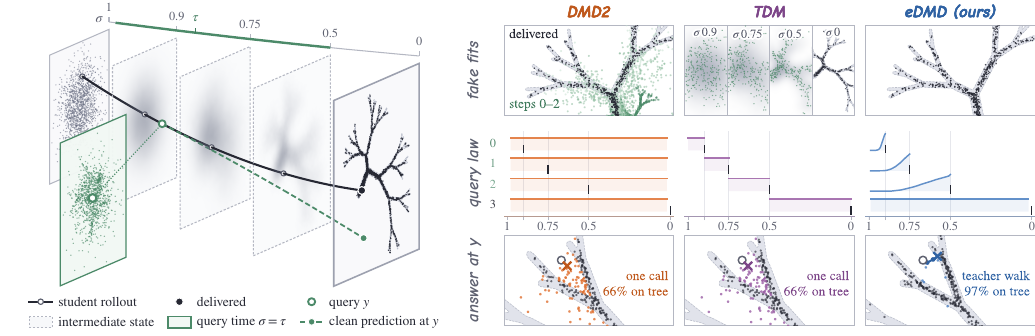}
\caption{\textbf{Overview of \ours on a 2D tree-structured toy distribution (gray).}
Left: a 4-step student denoises from noise ($\sigma=1$) to its final output ($\sigma=0$, ``delivered''); each update noises one step's prediction to a level $\tau$ (green bar: its range) and compares the teacher and the fake model at the resulting query $\y$.
Right: compared with DMD2 and TDM, \ours trains the fake model on final outputs only (top), draws the queries of intermediate steps from a calibrated density (middle; ticks: the level each step denoises to), and answers low-noise queries with a multi-step teacher instead of a single prediction (bottom; percentages: answers on the tree).}
\label{fig:overview}
\end{figure*}

Existing multi-step DMD methods incorporate intermediate predictions or states into the training of the fake model (\cref{fig:overview}), following two paradigms.
The first, represented by DMD2~\cite{yin2024dmd2}, trains a shared fake model using predictions collected from different denoising steps, effectively learning a mixture of intermediate and final prediction distributions.
The second, represented by TDM~\cite{luo2025tdm}, constructs stage-dependent references from intermediate denoising states and performs distribution matching within their corresponding noise intervals.
Despite their different implementations, both paradigms couple the distribution matching objective to intermediate distributions, whereas the ultimate goal of distillation is to match the student's final output distribution to the teacher's.
This distinction is particularly important for few-step students, which compress a complex generative process into only a few steps.
Their intermediate predictions need not follow the teacher's corresponding distributions, even when their final outputs accurately reproduce the teacher's distribution.
Therefore, imposing distributional constraints on intermediate predictions may introduce unnecessary restrictions and interfere with the optimization of final generation quality.

To understand the consequences of this mismatch, we theoretically analyze how the fake model's reference distribution influences the student update.
We denote the distribution of the student's final outputs as the \emph{endpoint distribution}.
Under exact conditional denoisers, we decompose the DMD correction signal into three components (\cref{eq:split}): a \emph{teacher enhancement} term arising from guidance, an \emph{endpoint matching} term that measures the discrepancy between the teacher and the student's final output distribution, and an additional \emph{reference residual} caused by the discrepancy between the fake model's reference distribution and the endpoint distribution.
Our analysis reveals that existing intermediate-based references can introduce a nonzero residual that fundamentally changes the intended distribution matching behavior.
Even when the student's final output distribution already matches the teacher's, this residual may remain nonzero and continue to push the student away from the desired distribution.
We further prove that the reference residual vanishes if and only if the fake model represents the endpoint distribution, regardless of the student's intermediate distributions (\cref{prop:reference}).
Our two-dimensional synthetic experiments confirm these effects, showing that intermediate-based references deflect the update from the desired direction, whereas an endpoint-aligned reference does not (\cref{fig:toy}).\looseness=-1

These findings motivate a simple principle: the fake model should represent the distribution of the samples that distillation ultimately aims to optimize, rather than the intermediate predictions used to generate them.
Based on this principle, we propose Endpoint Distribution Matching Distillation (\ours), which trains a single shared fake model exclusively on the student's final generated outputs and uses it to provide correction signals for every denoising step.
By aligning the fake model's reference with the endpoint distribution, \ours eliminates the reference residual and ensures that the distribution matching signal is determined by the discrepancy between the teacher and the student's final outputs, independently of the intermediate sampling trajectory (\cref{cor:endpoint}).
This allows intermediate steps to take whatever form best suits few-step generation, without imposing unnecessary distributional constraints.

Beyond correcting the student-side distribution reference, effective DMD also depends on the quality of the supervision provided by the teacher.
D-DMD~\cite{liu2026ddmd} identifies CFG augmentation of the teacher prediction as the primary engine of few-step distillation.
However, existing methods typically obtain teacher targets through a single guided network evaluation at each sampled noise level.
Such predictions can average over multiple plausible outputs and lose fine-grained details, whereas the teacher achieves high-quality generation through iterative denoising (\cref{fig:overview}, bottom).
To improve the precision of teacher supervision, we introduce a \emph{precise teacher (PT)} that replaces single-step teacher predictions with short multi-step guided trajectories within the final denoising interval, providing sharper and more accurate supervision without increasing student inference cost.
Furthermore, teacher predictions are not equally reliable across noise levels, as guidance and model approximation errors can introduce inconsistencies between predictions at different noise levels.
Uniformly sampling noise levels may therefore expose intermediate student steps to unreliable supervision.
To address this issue, we introduce a \emph{calibrated query density} that estimates the teacher's cross-level prediction consistency before distillation and concentrates intermediate-step queries on noise levels that provide more reliable supervision.

Overall, \ours aligns the distribution matching objective with the student's final outputs and improves the quality of teacher supervision.
We evaluate \ours on four representative diffusion and flow matching backbones (\sdthree~\cite{esser2024sd3}, \flux, \fluxnine~\cite{bfl2026flux2klein}, and \qwen-2.1~\cite{qwen2026qwenimage21}), distilling each into a 4-step generator without using real images.
Extensive experiments demonstrate that \ours consistently outperforms existing distribution matching baselines, achieving the highest peak \hpspp~\cite{liu2026hpsv3pp} scores across all four backbones.
Our main contributions are summarized as follows:
\begin{itemize}[leftmargin=*,itemsep=1pt,topsep=2pt]
\item We identify and theoretically characterize an objective mismatch in multi-step DMD: training the fake model on intermediate predictions introduces a reference residual that distorts distribution matching and vanishes if and only if the fake model learns the endpoint distribution.
\item We propose \ours, which trains a shared fake model exclusively on the student's final outputs and strengthens teacher supervision through a multi-step precise teacher (PT) and a calibrated query density.
\item Experiments on four backbones show consistent improvements over existing distribution matching methods, with higher human-preference scores, improved fine-grained details, and fewer structural artifacts.
\end{itemize}

\section{Related work}
\label{sec:related}

\noindent\textbf{Few-step diffusion distillation.}
Few-step distillation trains a student to match the quality of a diffusion or flow matching teacher~\cite{ho2020ddpm,song2021sde,lipman2023flowmatching} in only a few sampling steps.
Trajectory-based methods train the student to follow the teacher's sampling path: progressive distillation merges adjacent steps~\cite{salimans2022progressive,meng2023distillation}, rectified flows straighten the path~\cite{liu2023rectifiedflow,liu2024instaflow}, and consistency and flow-map methods learn direct mappings from noisy states to cleaner states~\cite{song2023consistency,luo2023lcm,kim2024ctm,wang2024pcm,ren2024hypersd,lu2025scm,sabour2025alignflow}.
Adversarial methods instead align output distributions using a discriminator against real or teacher samples~\cite{sauer2024add,sauer2024ladd,lin2024sdxllightning}.
Distribution matching, which our work builds upon, aligns them using score estimates from the teacher and an online fake model.

\noindent\textbf{Distribution matching distillation.}
Building on score distillation for text-to-3D generation~\cite{poole2023dreamfusion,wang2023prolificdreamer}, Diff-Instruct and DMD update the student with the score difference between the pretrained teacher and a fake model trained online on the student's samples~\cite{luo2023diffinstruct,yin2024dmd}.
Subsequent work explores alternative divergences~\cite{zhou2024sid,xu2025fdistill,wang2025uniinstruct}, incorporates adversarial objectives~\cite{yin2024dmd2,lu2025adm}, refines the fake score estimate or the update direction~\cite{fan2026phaseddmd,wu2026sgmd}, or mitigates mode collapse~\cite{wu2026dpdmd,li2026onexdistill}.
DMD2 extends DMD to multi-step generation~\cite{yin2024dmd2}, a setting adopted by many later methods~\cite{liu2026ddmd,liu2026cdm,luo2025tdm,zhou2025sidfewstep,liu2026crossdistill,wu2026dpdmd}.
Their fake models are typically trained on intermediate predictions or states, whereas we show that only a fake model trained on final outputs avoids biasing the student update.
This also explains a choice found in some implementations: ADM fits its fake model to final outputs in its training algorithm~\cite{lu2025adm}, and the final-step matching variant of SiD supervises only the final output by back-propagating through the entire sampling chain~\cite{zhou2025sidfewstep}, whereas \ours corrects every step directly.\looseness=-1

\noindent\textbf{Teacher supervision.}
Distillation relies on the quality of the teacher's supervision.
For diffusion models, generation quality often depends on guidance, from CFG~\cite{ho2022cfg} and its refinements~\cite{kynkaanniemi2024interval,wang2024cfgschedulers,sadat2025apg} to guidance by a weaker model~\cite{karras2024autoguidance} or by perturbed attention~\cite{ahn2024pag}.
D-DMD shows that the same holds in DMD: CFG augmentation of the teacher prediction is the primary engine of few-step distillation, with distribution matching acting as a regularizer~\cite{liu2026ddmd}.
Strengthening the teacher's supervision is therefore a direct way to improve the student, for example by fine-tuning the teacher~\cite{luo2025tdm}, targeting a reward-tilted teacher distribution~\cite{huang2026rtdmd}, or deriving regression targets from the teacher's own denoising outputs~\cite{zhou2023sparsefusion,sauer2024add,ge2026senseflow,wu2026dpdmd}.\looseness=-1

\makeatletter
\@ifundefined{corollary}{\theoremstyle{plain}\newtheorem{corollary}{Corollary}\theoremstyle{remark}}{}
\makeatother

\section{Reference distribution of the fake model}
\label{sec:prelim}

We now analyze how the distribution that the fake model is trained on, which we call its \emph{reference distribution}, shapes the student update in multi-step DMD.

\subsection{Multi-step distribution matching}
\label{sec:setup}

\noindent\textbf{Distribution matching.}
DMD trains a student to match a pretrained teacher's distribution~\cite{yin2024dmd}.
We first consider a single-step student that generates samples $\xh\sim q$ in $\mathbb R^n$, and write $p$ for the teacher distribution.
DMD minimizes the reverse KL divergence between their noise-perturbed distributions $q_\tau$ and $p_\tau$, where samples are perturbed as $\y=(1-\tau)\x+\tau\eps$ at level $\tau\in(0,1)$, with $\eps\sim\mathcal N(0,I_n)$~\cite{liu2023rectifiedflow,esser2024sd3}.
For a distribution $r$, let $D_r(\y,\tau)$ denote its denoiser, \ie, the prediction of a diffusion model trained on $r$: the average clean sample given the noisy sample $\y$.
The gradient with respect to the student parameters $\theta$ is
\begin{equation}
\begin{aligned}
&\nabla_\theta\,\mathrm{KL}\big(q_\tau\,\|\,p_\tau\big)\\
&\quad=-\frac{(1-\tau)^2}{\tau^2}\,\E\Big[\big(D_p(\y,\tau)-D_q(\y,\tau)\big)\,\frac{\partial\xh}{\partial\theta}\Big].
\end{aligned}
\label{eq:dmd}
\end{equation}
The student is thus moved along the difference between two denoisers: $D_p$ is given by the teacher, and $D_q$ is estimated by a fake model $F$ trained online on the student's samples.\looseness=-1

\noindent\textbf{Multi-step students.}
While one-step generation is efficient, a single network evaluation often struggles to preserve fine-grained details and complex structures.
Multi-step students address this limitation by progressively refining their predictions over a few denoising steps~\cite{yin2024dmd2}.
A $K$-step student $G_\theta$ starts from pure noise $\z_0$ and follows a noise grid $1=\sigma_0>\dots>\sigma_K=0$.
At step $k=0,\dots,K-1$, it moves from state $\z_k$ to $\z_{k+1}$ and provides a \emph{clean prediction} $\xh_k$, an estimate of the clean sample given $\z_k$.
At the last step, this prediction coincides with the final output $\z_K$.
We denote the prediction distributions by $q_k$ and the final output distribution by $q$, the \emph{endpoint distribution}.

\noindent\textbf{Multi-step update.}
During multi-step DMD training, we select a step $k$ and construct a noisy input $\y$ from its clean prediction $\xh_k$ or its resulting state $\z_{k+1}$.
This input and its noise level $\tau$ form a \emph{query} to the teacher and fake model.
Denoting their predictions by $A(\y,\tau)$ and $F(\y,\tau)$, respectively, we use the difference $C=A-F$ to correct $\xh_k$ through the surrogate loss~\cite{yin2024dmd,yin2024dmd2}
\begin{equation}
\mathcal L_G=\E\,\big\|\xh_k-\sg\big[\xh_k+C/\omega\big]\big\|^2,
\label{eq:update}
\end{equation}
where $\sg$ stops gradients and $\omega=\|\xh_k-A\|_1/n$ is the mean absolute difference over the sample's $n$ coordinates.
The teacher target $A$ is often obtained by enhancing $D_p$ with guidance such as CFG~\cite{ho2022cfg}.
The fake model $F$ is trained online on selected student samples.
We denote their distribution by $\nu$ and call it the \emph{reference distribution}.

\noindent\textbf{Decomposing the correction.}
To understand how the fake model's training distribution affects this update, assume an exact fake model, $F=D_\nu$.
Adding and subtracting the teacher denoiser $D_p$ and the endpoint denoiser $D_q$ gives
\begin{equation}
C=\underbrace{(A-D_p)}_{\text{teacher enhancement}}+\underbrace{(D_p-D_q)}_{\text{endpoint matching}}+\underbrace{(D_q-D_\nu)}_{\text{reference residual }R}.
\label{eq:split}
\end{equation}
The \emph{teacher enhancement} strengthens teacher supervision, for example through CFG~\cite{liu2026ddmd}.
The \emph{endpoint matching} term is the distribution matching direction for the student's final outputs, as in \cref{eq:dmd}.
The \emph{reference residual} $R$ is the extra correction caused by a mismatch between the fake reference and the student's endpoint distribution.

\begin{figure}[t]
\centering
\includegraphics[width=\linewidth]{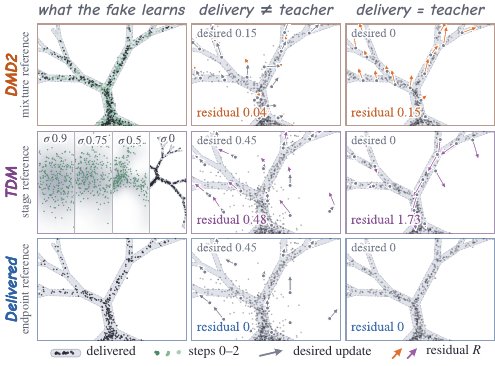}
\caption{\textbf{Intermediate-based references deflect the update from the desired direction; the endpoint reference does not.}
Each row shows one reference (mixture, stage, endpoint): what the fake model learns (left) and one update of the final outputs for a student with mismatched ($q\neq p$, middle) and matched ($q=p$, right) final outputs.
Gray arrows show the desired update, \ie, the endpoint matching term $D_p-D_q$ of \cref{eq:split}, which is zero for the matched student; colored arrows show the additional displacement caused by the reference residual (numbers: mean norms).}
\label{fig:toy}
\end{figure}

\subsection{The reference residual}
\label{sec:reference}

\begin{proposition}[Reference residual]
\label{prop:reference}
Assume exact conditional denoisers under a shared forward noising process.
At every noise level $0<\tau<1$,
\begin{equation}
R=D_q-D_\nu=\frac{\tau^2}{1-\tau}\,\nabla_\y\log\frac{q_\tau}{\nu_\tau},
\end{equation}
and $R(\cdot,\tau)\equiv0$ if and only if $\nu=q$.
\end{proposition}
\noindent\textbf{Mixture reference.}
DMD2 and D-DMD train a single fake model on the clean prediction of a randomly selected student step~\cite{yin2024dmd2,liu2026ddmd}.
Let $J$ denote the selected step index, and $\pi_k=P(J=k)$ the probability of selecting step $k$.
The fake model thus learns the mixture $\nu=\sum_k\pi_kq_k$.
CDM similarly trains on a mixture over Euler rollouts of random lengths and schedules~\cite{liu2026cdm}.
By \cref{prop:reference} and Bayes' rule, the residual at the final step is
\begin{equation}
R_{\rm mix}(\y,\tau)=\frac{\tau^2}{1-\tau}\,\nabla_\y\log P(J=K-1\mid\Y_\tau=\y),
\label{eq:rmix}
\end{equation}
where $P(J=K-1\mid\Y_\tau=\y)$ is the probability that a noisy sample $\y$ comes from the final step; the residual thus pushes final outputs toward regions where the last step's density exceeds that of earlier steps (App.~\ref{app:proofs-mix}).

\noindent\textbf{Stage reference.}
TDM noises each state $\z_{k+1}$ within its stage interval and matches its distribution $r_k$ to the teacher marginal $p_{\sigma_{k+1}}$ via a shared fake model~\cite{luo2025tdm}.
The final stage coincides with the endpoint reference, but every earlier stage adds a correction whenever $r_k\neq p_{\sigma_{k+1}}$ (App.~\ref{app:proofs-sets}).

The following result summarizes when each reference stops correcting the student.
\begin{proposition}[Zero-correction sets]
\label{prop:zerosets}
Let $A=D_p$ and the denoisers be exact.
The students whose correction vanishes across all queries and steps are
\begin{equation}
\begin{aligned}
&\text{endpoint:}\quad &&\{q=p\},\\
&\text{mixture:}\quad &&\{\textstyle\sum_k\pi_kq_k=p\},\\
&\text{stage:}\quad &&\{q=p,\ r_k=p_{\sigma_{k+1}}\ \forall k<K-1\}.
\end{aligned}
\end{equation}
\end{proposition}
\noindent Only the endpoint reference stops exactly at $q=p$.
The mixture reference also stops at students with $q\neq p$ whose endpoint errors are offset by earlier steps, whereas the stage reference keeps correcting students with $q=p$ whose intermediate states differ from the teacher's.

\subsection{Verification on a 2D synthetic distribution}
\label{sec:toy}

\Cref{fig:toy} tests the three references on a 2D tree-structured distribution (App.~\ref{app:toy}).
We use the unguided target $A=D_p$, so that the correction reduces to the endpoint matching term, which is the desired update, and the residual (\cref{eq:split}).
We consider two students whose intermediate steps deviate from the teacher's, one with mismatched final outputs ($q\neq p$) and one with matched final outputs ($q=p$).
For the mismatched student, all three references contain the desired update, but the mixture and stage references add a residual that deflects it.
For the matched student, the desired update is zero and the endpoint reference yields no correction, whereas the mixture and stage references still push the student away from the teacher.
Even students producing identical final samples can receive nearly opposite residual updates when their intermediate distributions differ (\cref{fig:app-toy-students}).
In full training, the endpoint reference converges closest to $p$ and preserves an already converged student (\cref{fig:app-toy-train}).

\section{Endpoint distribution matching}
\label{sec:method}

Building on the analysis in \cref{sec:prelim}, \ours trains its fake model only on the student's final outputs to remove the reference residual.
This lets us improve teacher supervision without interference from intermediate-based references.
We do so by refining low-noise targets with a precise teacher (PT) and calibrating the query density to favor more reliable predictions.

\subsection{Endpoint reference}
\label{sec:endpoint}

\ours trains the fake model solely on final outputs $\x\sim q$:
\begin{equation}
\mathcal L_F=\E\Big[\lambda(\tau)\,\big\|F\big((1-\tau)\x+\tau\eps,\tau\big)-\x\big\|^2\Big],
\label{eq:critic}
\end{equation}
where $\eps\sim\mathcal N(0,I_n)$, $\tau\sim\mathcal U(0,1)$, and $\lambda(\tau)$ is a clipped SNR weight (App.~\ref{app:impl}).
At the optimum $F=D_q$, the reference residual in \cref{eq:split} vanishes for any teacher target $A$, leaving $C=A-D_q$ at every step.

\begin{corollary}[Endpoint reference]
\label{cor:endpoint}
Let $\nu=q$.
If $A=D_{p^\star}$ for a target distribution $p^\star$, the correction vanishes across all steps and queries if and only if $q=p^\star$, regardless of intermediate distributions or the query density.
\end{corollary}

\noindent\textbf{From the correction to the update.}
For each sample, we select a student step $k$ with probability $\pi_k$ and form its query $\y_k$ at level $\tau$.
The rollout is detached, so gradients flow only through the selected step's clean prediction $\xh_k$.
Applying \cref{eq:update} gives the expected parameter update
\begin{equation}
\Delta\theta\ \propto\ \sum_k\pi_k\,\E\bigg[\Big(\frac{\partial\xh_k}{\partial\theta}\Big)^{\!\top}\frac{A(\y_k,\tau)-D_q(\y_k,\tau)}{\omega_k}\bigg].
\label{eq:dtheta}
\end{equation}

\subsection{A better teacher at query time}
\label{sec:teacher}
\begin{table*}[t]
\centering
\footnotesize
\tabcolsep=4pt
\caption{\textbf{Comparison of 4-step students distilled from \sdthree, \flux, and \fluxnine.}
All metrics are evaluated on DrawBench-1000 except DPG, which uses DPG-Bench.
Peak: highest \hpspp within 2{,}000 updates over raw/EMA weights (\cref{tab:app-ckpt}); @1k/@2k: raw \hpspp after 1k/2k updates; `--': unavailable.
$^\ast$: official checkpoint; other baselines: trained by us.
CLIP: ViT-B/32~\cite{radford2021clip}.
\textbf{Bold}/\underline{underline}: best/second-best student; gray: unranked references.}
\label{tab:main}
\begin{tabular}{@{}l *{10}{c}@{}}
\toprule
 & \multicolumn{8}{c}{Scores at the peak checkpoint} & \multicolumn{2}{c}{Fixed steps} \\
\cmidrule(lr){2-9}\cmidrule(lr){10-11}
Method & \hpspp$\uparrow$ & HPSv3$\uparrow$ & PickScore$\uparrow$ & Aesthetic$\uparrow$ & HPSv2.1$\uparrow$ & ImageReward$\uparrow$ & CLIP$\uparrow$ & DPG$\uparrow$ & @1k$\uparrow$ & @2k$\uparrow$ \\
\midrule
\multicolumn{11}{l}{\emph{\sdthree, teacher default $w=7$}} \\
\dashedmidrule{11}
\rowcolor{black!5}
Teacher (28 steps) & 8.554 & 9.257 & 22.429 & 5.350 & 0.2849 & 0.964 & 0.3263 & 85.93 & -- & -- \\
DMD2~\cite{yin2024dmd2} & 10.270 & 10.129 & 22.455 & \underline{5.547} & \underline{0.3052} & \textbf{1.136} & 0.3210 & \underline{84.77} & 10.209 & 9.354 \\
TDM$^\ast$~\cite{luo2025tdm} & 10.386 & 10.418 & 22.495 & 5.499 & 0.2996 & 1.016 & 0.3183 & 83.59 & -- & -- \\
D-DMD~\cite{liu2026ddmd} & \underline{10.841} & 11.161 & 22.587 & \textbf{5.563} & 0.3041 & 1.043 & 0.3203 & 84.49 & \underline{10.388} & \underline{10.677} \\
CDM$^\ast$~\cite{liu2026cdm} & 10.396 & \textbf{11.890} & \textbf{22.622} & 5.490 & 0.2984 & 0.994 & \textbf{0.3234} & 84.56 & -- & -- \\
\rowcolor{oursrow}
\ours & \textbf{11.171} & \underline{11.336} & \underline{22.589} & \underline{5.547} & \textbf{0.3097} & \underline{1.115} & \underline{0.3211} & \textbf{85.33} & \textbf{11.171} & \textbf{10.820} \\
\midrule
\multicolumn{11}{l}{\emph{\flux, teacher default $w=4$}} \\
\dashedmidrule{11}
\rowcolor{black!5}
Teacher (50 steps) & 9.236 & 9.255 & 22.360 & 5.371 & 0.2760 & 0.832 & 0.3166 & 83.12 & -- & -- \\
\rowcolor{black!5}
Distilled (4 steps)$^\ast$ & 10.509 & 10.586 & 22.815 & 5.352 & 0.2904 & 1.057 & 0.3182 & 85.79 & -- & -- \\
DMD2~\cite{yin2024dmd2} & \underline{11.337} & \underline{10.626} & 22.302 & 5.582 & \underline{0.3004} & 1.025 & 0.3076 & \underline{83.22} & \underline{10.966} & \underline{11.055} \\
TDM~\cite{luo2025tdm} & 10.259 & 9.635 & 22.268 & 5.477 & 0.2857 & 0.961 & \textbf{0.3119} & 82.09 & 10.075 & 10.254 \\
D-DMD~\cite{liu2026ddmd} & 10.228 & 9.803 & 22.308 & 5.428 & 0.2828 & 0.939 & \underline{0.3100} & 81.36 & 9.793 & 9.495 \\
CDM~\cite{liu2026cdm} & 10.790 & 10.350 & \underline{22.500} & \textbf{5.658} & 0.2995 & \underline{1.033} & 0.3062 & 83.19 & 9.445 & 9.185 \\
\rowcolor{oursrow}
\ours & \textbf{12.699} & \textbf{11.829} & \textbf{22.721} & \underline{5.600} & \textbf{0.3042} & \textbf{1.129} & 0.3099 & \textbf{83.41} & \textbf{12.072} & \textbf{12.699} \\
\midrule
\multicolumn{11}{l}{\emph{\fluxnine, teacher default $w=4$}} \\
\dashedmidrule{11}
\rowcolor{black!5}
Teacher (50 steps) & 9.785 & 9.648 & 22.673 & 5.290 & 0.2815 & 0.979 & 0.3266 & 85.68 & -- & -- \\
\rowcolor{black!5}
Distilled (4 steps)$^\ast$ & 11.526 & 10.954 & 23.062 & 5.400 & 0.2992 & 1.188 & 0.3249 & 86.37 & -- & -- \\
DMD2~\cite{yin2024dmd2} & 11.104 & 9.761 & 22.652 & \underline{5.505} & 0.3007 & 1.168 & 0.3183 & 82.93 & 11.038 & 10.416 \\
TDM~\cite{luo2025tdm} & 11.084 & 9.988 & 22.585 & 5.424 & 0.2979 & 1.135 & \underline{0.3223} & 84.36 & \underline{11.062} & 11.004 \\
D-DMD~\cite{liu2026ddmd} & \underline{11.325} & 10.076 & 22.674 & 5.490 & \underline{0.3015} & \underline{1.182} & 0.3198 & 83.84 & 11.027 & \underline{11.135} \\
CDM~\cite{liu2026cdm} & 11.077 & \underline{10.292} & \underline{22.799} & \textbf{5.540} & 0.2987 & 1.137 & 0.3183 & \underline{84.38} & 8.239 & 10.587 \\
\rowcolor{oursrow}
\ours & \textbf{12.431} & \textbf{11.493} & \textbf{22.984} & 5.458 & \textbf{0.3085} & \textbf{1.242} & \textbf{0.3260} & \textbf{86.41} & \textbf{11.887} & \textbf{12.283} \\
\bottomrule
\end{tabular}

\end{table*}

\begin{figure*}[t]
\centering
\includegraphics[width=0.94\linewidth]{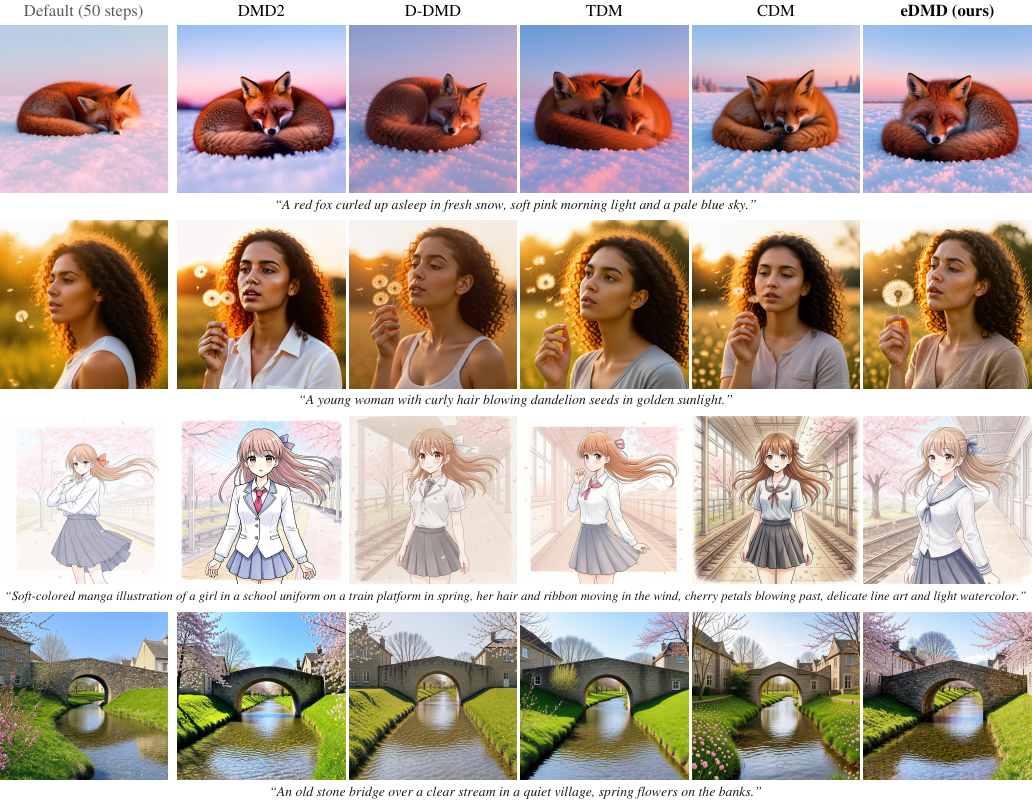}
\vspace{-2mm}
\caption{\textbf{Qualitative results on \flux.} Leftmost: teacher (50 steps, $w=4$); others: 4-step students at the checkpoints of \cref{tab:main}; shared prompt and noise per row.
Best viewed zoomed in: on fine structures such as fur, hair, and grass, \ours preserves richer details and fewer artifacts, yielding more natural textures than existing methods.}
\label{fig:qual}
\vspace{-3mm}
\end{figure*}

With the unguided prediction $A=D_p$, the student is matched to the teacher's unguided distribution~\cite{liu2026cdm}.
We strengthen teacher supervision with guidance~\cite{liu2026ddmd} and refine its predictions through short multi-step integration.

\noindent\textbf{Guidance.}
With guidance strength $\gamma$, we extrapolate a conditional prediction $T$ away from a weaker prediction $T^-$:
\begin{equation}
B=T+\gamma\,(T-T^-),
\label{eq:contrast}
\end{equation}
For CFG, $T=\Tc=D_p$ and $T^-=\Tu$ are the conditional and unconditional predictions~\cite{ho2022cfg}.
The guided target is $B=\Tw=\Tu+w(\Tc-\Tu)$, where $w=1+\gamma$ is the usual CFG scale.
PAG obtains $T^-$ by perturbing self-attention~\cite{ahn2024pag}; autoguidance uses a weaker model~\cite{karras2024autoguidance}.

\noindent\textbf{Precise teacher.}
\Cref{fig:overview} (bottom) illustrates a limitation of single-step teacher supervision: a prediction can fall between branches of the tree, missing the fine structure of the data distribution.
Our precise teacher (PT) follows a short denoising trajectory from the same query, yielding targets that better follow the tree.
Concretely, we integrate the teacher's sampling ODE driven by $B$ from the query level $\tau$ to $0$ and use the resulting clean sample as the target.
Since the final step directly produces the output sample, we focus PT on its low-noise queries ($\tau<\sigma_{K-1}$) to refine the final details, retaining the single-step target elsewhere:
\begin{equation}
A(\y,\tau)=\begin{cases}\Phi^B_{\tau\to0}(\y), & k=K-1,\ \tau<\sigma_{K-1},\\ B(\y,\tau), & \text{otherwise},\end{cases}
\label{eq:pt}
\end{equation}
where $\Phi^B_{\tau\to0}$ denotes multi-step integration (App.~\ref{app:impl}).

\begin{algorithm}[t]
\caption{\textbf{\ours training.}}
\label{alg:etdm}
\small
\renewcommand{\algorithmicrequire}{\textbf{Input:}}
\begin{algorithmic}[1]
\REQUIRE guided teacher prediction $B$, grid $\sigma_0{=}1>\dots>\sigma_K{=}0$
\STATE student $G_\theta$ and fake $F$, both initialized from the teacher
\REPEAT
  \item[] \textcolor{commentcolor}{\textit{// generate a sample in $K$ steps, without gradient}}
  \STATE sample a prompt and $\z_0\sim\mathcal N(0,I)$
  \STATE $\hat{\mathbf v}_k\leftarrow\sg[G_\theta(\z_k,\sigma_k)]$ \hfill\textcolor{commentcolor}{\textit{// the step's velocity output}}
  \STATE $\z_{k+1}\leftarrow\z_k+(\sigma_{k+1}-\sigma_k)\,\hat{\mathbf v}_k$ for $k<K$ \hfill\cref{eq:euler}
  \STATE $\x\leftarrow\z_K$ \hfill\textcolor{commentcolor}{\textit{// the generated sample, at $\sigma_K=0$}}
  \item[] \textcolor{commentcolor}{\textit{// train the fake to denoise generated samples}}
  \STATE $\y'\leftarrow(1-\tau')\,\x+\tau'\eps'$ with $\tau'\sim\mathcal U(0,1)$
  \STATE $F\leftarrow F-\eta_F\nabla_F\mathcal L_F(\y',\tau')$ \hfill\cref{eq:critic}
  \item[] \textcolor{commentcolor}{\textit{// correct one student step at one query level}}
  \STATE pick a step $k$ and a query level $\tau\sim\rho_k$ \hfill\cref{eq:law}
  \STATE $\xh_k\leftarrow\z_k-\sigma_kG_\theta(\z_k,\sigma_k)$ \hfill\textcolor{commentcolor}{\textit{// with gradient}}
  \STATE form query $\y$ by noising $\xh_k$ to level $\tau$ \hfill\textcolor{commentcolor}{\textit{// query at $\tau$}}
  \STATE $A\leftarrow B(\y,\tau)$, the teacher's guided prediction \hfill\cref{eq:contrast}
  \item[] \hspace{1em}or PT, if $k=K-1$ and $\tau<\sigma_{K-1}$ \hfill\cref{eq:pt}
  \STATE $C\leftarrow A-F(\y,\tau)$ \hfill\textcolor{commentcolor}{\textit{// the correction}}
  \STATE $\theta\leftarrow\theta-\eta\,\nabla_\theta\mathcal L_G$, moving $\xh_k$ along $C$ \hfill\cref{eq:update}
\UNTIL{converged}
\end{algorithmic}
\end{algorithm}

\subsection{Calibrated query levels for intermediate steps}
\label{sec:query}

\noindent\textbf{Query range.}
We use a deterministic Euler rollout~\cite{luo2025tdm}, with velocity $\hat{\mathbf v}_k=G_\theta(\z_k,\sigma_k)$ and clean prediction $\xh_k=\z_k-\sigma_k\hat{\mathbf v}_k$:
\begin{equation}
\z_{k+1}=\z_k+(\sigma_{k+1}-\sigma_k)\,\hat{\mathbf v}_k.
\label{eq:euler}
\end{equation}
Step $k$ reaches noise level $s=\sigma_{k+1}$, leaving finer denoising to later steps; we therefore query it only at $\tau\ge s$~\cite{fan2026phaseddmd,huang2026rtdmd}.

\noindent\textbf{Query density.}
Intermediate steps receive local corrections that must support the remaining denoising process.
Uniform sampling gives every level in this range the same weight.
Yet guidance and model errors can make the teacher's predictions drift across noise levels, so a student step can receive inconsistent supervision depending on where the teacher is queried.
We favor query levels whose teacher predictions are more consistent with those at $s$ (\cref{fig:overview}, middle).
To assess this consistency, we compare predictions at $\tau$ and $s$ on teacher samples before distillation.
Adding noise naturally changes a prediction by removing information; our score estimates drift beyond this expected averaging.
The resulting normalized score $\hat b_s(\tau)$ (App.~\ref{app:calib}) defines the sampling density:
\begin{equation}
\tau\sim\rho_k\propto\exp\big(-\hat b_{\sigma_{k+1}}(\tau)\big)\quad\text{on }[\sigma_{k+1},1).
\label{eq:law}
\end{equation}

\subsection{Update schedule and training cost}
\label{sec:impl}

Following \cref{alg:etdm}, we perform five fake updates (each with fresh prompts and a detached rollout) per student update, which reuses the final rollout and backpropagates only through the selected step.
Query calibration is performed once per teacher and grid before training; sampling from the precomputed densities adds no teacher evaluations during distillation.
PT replaces a single guided teacher prediction with a short multi-step integration for final-step queries below $\sigma_{K-1}$.
These extra teacher evaluations are confined to training and do not increase student inference cost.
App.~\ref{app:impl} gives training settings and timing, and \cref{fig:app-time} compares methods under matched wall-clock time.

\section{Experiments}
\label{sec:exp}

\subsection{Setup}
\label{sec:expsetup}

\noindent\textbf{Teachers and training.}
We distill four teachers into 4-step students without using real images: \sdthree~\cite{esser2024sd3}, \flux and \fluxnine~\cite{bfl2026flux2klein}, and \qwen-2.1~\cite{qwen2026qwenimage21}.
During distillation, we use each teacher's default CFG scale ($w=7$ on \sdthree, $w=4$ on the FLUX.2 models, and $w=1$ on \qwen).
On \qwen, \ours uses PAG to enhance the teacher predictions (App.~\ref{app:qwen}).
The \sdthree and FLUX.2 runs train on the 191{,}500 prompts of CDM~\cite{liu2026cdm} for 2{,}000 updates on NPUs; \qwen trains on prompt-enhanced rewrites of these prompts for 600 updates on H800 GPUs (App.~\ref{app:impl}).

\noindent\textbf{Baselines.}
We compare against DMD2~\cite{yin2024dmd2}, D-DMD~\cite{liu2026ddmd}, TDM~\cite{luo2025tdm}, and CDM~\cite{liu2026cdm}, and include the official 4-step FLUX.2-klein models and 8-step \qwen-2.1-Turbo as unranked references.
DMD2 (without its GAN loss) and D-DMD are trained in our codebase using their released recipes, with learning rates adjusted for \sdthree and both FLUX.2 models (App.~\ref{app:baselines}).
On \sdthree, we evaluate the official TDM and CDM checkpoints with their native samplers; on the FLUX.2 models, we retrain CDM and use our $\ell_2$ re-implementation of TDM, which outperforms Huber loss on \flux.
On \qwen, we compare with DMD2 and TDM.

\noindent\textbf{Evaluation.}
We generate 1{,}000 images from the 200 DrawBench prompts~\cite{saharia2022imagen} using five seeds per prompt, forming DrawBench-1000.
We report \hpspp~\cite{liu2026hpsv3pp}, HPSv3~\cite{ma2025hpsv3}, PickScore~\cite{kirstain2023pickapic}, LAION Aesthetic~\cite{schuhmann2022laionaesthetics}, HPSv2.1~\cite{wu2023hpsv2}, ImageReward~\cite{xu2023imagereward}, CLIP score~\cite{radford2021clip}, and DPG-Bench~\cite{hu2024ella}.
For each run, we select the peak checkpoint by the highest \hpspp over raw and EMA weights evaluated every 100 updates.
For \sdthree and the FLUX.2 models, we also report raw \hpspp at 1k and 2k updates (multi-seed and held-out checks in App.~\ref{app:curves}).

\subsection{Main results}
\label{sec:main}

\noindent\textbf{Quantitative comparison.}
\Cref{tab:main} compares 4-step students on \sdthree and both FLUX.2 models.
\ours achieves the highest \hpspp at both the peak checkpoint and the fixed training budgets across all three teachers.
It also leads on DPG-Bench and HPSv2.1 across all three, and on HPSv3, PickScore, and ImageReward on both FLUX.2 models.
On \qwen-2.1 (\cref{tab:qwen}), \ours leads the 4-step students on all seven metrics.

\begin{table}[t]
\centering
\footnotesize
\tabcolsep=2.4pt
\caption{\textbf{4-step students on \qwen-2.1.}
DrawBench-1000 uses prompt-enhanced rewrites scored against original prompts; all methods use $w=1$.
Peak selection as in \cref{tab:main}, within 600 updates.
IR: ImageReward.
Gray: unranked teacher and official 8-step Turbo$^\ast$.
\textbf{Bold}/\underline{underline}: best/second-best student.}
\label{tab:qwen}
\resizebox{\linewidth}{!}{%
\begin{tabular}{@{}l *{7}{c}@{}}
\toprule
Method & \shortstack{HPS\\v3++$\uparrow$} & \shortstack{HPS\\v3$\uparrow$} & Pick$\uparrow$ & Aes.$\uparrow$ & \shortstack{HPS\\v2.1$\uparrow$} & IR$\uparrow$ & CLIP$\uparrow$ \\
\midrule
\rowcolor{black!5}
Teacher (40 steps) & 13.767 & 12.493 & 23.097 & 5.671 & 0.2974 & 1.057 & 0.3144 \\
\rowcolor{black!5}
Turbo$^\ast$ (8 steps) & 13.850 & 12.171 & 23.023 & 5.675 & 0.2985 & 1.064 & 0.3164 \\
DMD2 & \underline{14.212} & \underline{13.073} & 23.090 & \underline{5.781} & \underline{0.3032} & 1.049 & 0.3126 \\
TDM & 13.787 & 12.804 & \underline{23.101} & 5.743 & 0.3003 & \underline{1.072} & \underline{0.3129} \\
\rowcolor{oursrow}
\ours & \textbf{14.691} & \textbf{13.186} & \textbf{23.175} & \textbf{5.788} & \textbf{0.3109} & \textbf{1.104} & \textbf{0.3138} \\
\bottomrule
\end{tabular}}
\vspace{-3mm}
\end{table}

\noindent\textbf{Qualitative comparison.}\phantomsection\label{sec:qual}
\Cref{fig:qual} compares the distilled students on \flux using shared prompts and initial noise, alongside the teacher's default sampler.
\ours preserves fine details in fur, hair, and grass.
Additional qualitative comparisons across all teachers are shown in App.~\ref{app:qual}.
\begin{table}[t]
\centering
\footnotesize
\caption{\textbf{Component ablations trained from scratch.}
E: endpoint reference; PT: precise teacher; Q: calibrated query density.
Peak selection as in \cref{tab:main}; IR: ImageReward.}
\label{tab:ablation}
\newcommand{\ablrows}{%
\toprule
Variant & \shortstack{HPS\\v3++$\uparrow$} & \shortstack{HPS\\v3$\uparrow$} & Pick$\uparrow$ & Aes.$\uparrow$ & \shortstack{HPS\\v2.1$\uparrow$} & IR$\uparrow$ \\
\midrule
\multicolumn{7}{@{}l}{\emph{\sdthree}} \\
\dashedmidrule{7}
\rowcolor{oursrow}
Full & 11.171 & 11.336 & 22.589 & 5.547 & 0.3097 & 1.115 \\
w/o E & 10.122 & 11.166 & 22.480 & 5.500 & 0.3051 & 1.056 \\
w/o PT & 11.026 & 11.092 & 22.547 & 5.497 & 0.3096 & 1.115 \\
w/o Q & 10.798 & 10.706 & 22.562 & 5.490 & 0.3063 & 1.110 \\
\midrule
\multicolumn{7}{@{}l}{\emph{\flux}} \\
\dashedmidrule{7}
\rowcolor{oursrow}
Full & 12.699 & 11.829 & 22.721 & 5.600 & 0.3042 & 1.129 \\
w/o E & 11.095 & 10.637 & 22.496 & 5.487 & 0.2878 & 0.950 \\
w/o PT & 11.667 & 10.627 & 22.465 & 5.573 & 0.3004 & 1.089 \\
w/o Q & 12.507 & 11.592 & 22.692 & 5.607 & 0.3019 & 1.113 \\
\bottomrule}
\ifdefined\ablnat\else\newlength{\ablnat}\fi
\settowidth{\ablnat}{{\tabcolsep=0pt\begin{tabular}{@{}l *{6}{c}@{}}\ablrows\end{tabular}}}
\setlength{\tabcolsep}{\dimexpr(\linewidth-\ablnat)/12\relax}
\begin{tabular}{@{}>{\columncolor{white}[0pt][\tabcolsep]}l *{5}{c} >{\columncolor{white}[\tabcolsep][0pt]}c@{}}\ablrows\end{tabular}

\end{table}

\subsection{Ablations and analysis}
\label{sec:ablation}

\noindent\textbf{Protocol.}
\Cref{tab:ablation} ablates one component at a time from scratch (training curves in \cref{fig:app-curves}).
Without E, the fake model is trained on intermediate landing states with TDM's weighting (\cref{tab:app-tdmfake}); without PT, all queries use a single-step guided teacher prediction; without Q, intermediate steps are queried uniformly on $[\sigma_{k+1},0.99)$.
For reference, independent reruns of the same configuration vary by a standard deviation of 0.074 on \sdthree and 0.11 on \flux across matched checkpoints (App.~\ref{app:curves}).

\noindent\textbf{Endpoint reference.}
Removing E causes the largest overall degradation among the component ablations on both \sdthree and \flux (\cref{tab:ablation}), with persistent gaps during later training (\cref{fig:app-curves}).
To test whether the benefit extends beyond \ours, we train TDM with an endpoint fake while retaining its student objective and rollout.
This improves both peak and fixed-budget results on \sdthree and both FLUX.2 models (App.~\ref{app:plugin}).
We also switch a trained \flux student to the mixture reference, keeping all other settings fixed.
Sample quality declines (\cref{fig:app-fork}), and earlier clean predictions move closer to the final output, consistent with \cref{sec:reference}.

\noindent\textbf{Precise teacher.}
The PT ablation tests the value of replacing a single teacher prediction with a short denoising trajectory.
Multi-step targets improve generation quality on both teachers (\cref{tab:ablation}), with gains concentrated early in training on \sdthree and a sustained improvement on \flux (\cref{fig:app-curves}).

\noindent\textbf{Query density.}
Query calibration uses the same rule for every teacher, with the density determined before student training from cross-level prediction consistency.
It improves generation quality on \sdthree, where stronger prediction drift leads the calibration to favor queries nearer the landing level.
On \flux, the calibrated density remains nearly uniform, and the two sampling schemes achieve similar performance (\cref{tab:ablation}; App.~\ref{app:calib}).

\section{Conclusion}
\label{sec:concl}

We showed that training the fake model on intermediate predictions or states can distort distribution matching in multi-step DMD.
The resulting reference residual can persist even after the student's final output distribution matches the teacher's.
\ours resolves this mismatch with an endpoint reference, training a shared fake model exclusively on final outputs and using it to supervise every denoising step.
This keeps distribution matching focused on the samples delivered at inference and gives intermediate predictions the freedom to adapt to few-step generation.
We further improve teacher supervision in two ways: a precise teacher (PT) refines low-noise targets through multi-step denoising, and a calibrated query density favors noise levels where teacher predictions are more reliable.
Experiments on four diffusion and flow matching backbones show that \ours improves 4-step generation quality over existing distribution matching methods.

{
    \small
    \bibliographystyle{ieeenat_fullname}
    \bibliography{main}
}

\clearpage
\setcounter{page}{1}
\ifarxiv
\pdfbookmark[0]{Appendix}{appendix.contents}
\addtocontents{toc}{\protect\setcounter{tocdepth}{2}}
\begingroup
\small
\noindent\begin{minipage}{\linewidth}
{\centering\large\bfseries Appendix\par}
\vspace{5pt}
\setcounter{tocdepth}{2}
\appendixcontents
\end{minipage}\par
\endgroup
\medskip
\else
\maketitlesupplementary
\fi
\appendix
\setcounter{equation}{0}
\setcounter{figure}{0}
\setcounter{table}{0}
\renewcommand{\theequation}{A.\arabic{equation}}
\renewcommand{\thefigure}{A.\arabic{figure}}
\renewcommand{\thetable}{A.\arabic{table}}
\renewcommand{\theHequation}{appendix.\arabic{equation}}
\renewcommand{\theHfigure}{appendix.\arabic{figure}}
\renewcommand{\theHtable}{appendix.\arabic{table}}
\crefalias{section}{appendix}
\crefalias{subsection}{appendix}
\makeatletter
\@ifundefined{corollary}{\theoremstyle{plain}\newtheorem{corollary}{Corollary}\theoremstyle{remark}}{}
\makeatother

\section{Proofs}
\label{app:proofs}

\noindent\textbf{Setting.}
Notation follows \cref{sec:setup}.
We fix a prompt and $0<\tau<1$, assume finite second moments for clean distributions, and use exact denoisers.
For a clean distribution $r$, the noised density at level $\tau$ is
\begin{equation}
r_\tau(\y)=\int\varphi_\tau\big(\y-(1-\tau)\x\big)\,r(\mathrm d\x),
\label{eq:app-noised}
\end{equation}
with $\varphi_\tau$ the density of $\mathcal N(0,\tau^2I_n)$.
It is positive and extends to an entire function of $\y$, so it is real-analytic on $\mathbb R^n$.
Tweedie's formula~\cite{efron2011tweedie} gives the denoiser
\begin{equation}
D_r(\y,\tau)=\frac{\y+\tau^2\,\nabla\log r_\tau(\y)}{1-\tau}.
\label{eq:app-tweedie}
\end{equation}
For $0\le s\le\tau<1$, set $a=(1-\tau)/(1-s)$.
The forward kernel $P_{s\to\tau}$ maps $\z$ to
\begin{equation}
P_{s\to\tau}:\ \z\mapsto a\,\z+\sqrt{\tau^2-a^2s^2}\;\bm\zeta,
\label{eq:app-kernel}
\end{equation}
with $\bm\zeta\sim\mathcal N(0,I_n)$.
Applied to an observation noised to level $s$, this kernel gives an observation noised to level $\tau$.
At a level with nonzero added Gaussian noise, the query density is positive on $\mathbb R^n$.
By continuity, almost-sure zero correction then means that the field is identically zero.

\subsection{The reference residual}
\label{app:proofs-ref}

\noindent\textbf{Proof of \cref{prop:reference}.}
Subtracting \cref{eq:app-tweedie} for $\nu$ and $q$ gives
\begin{equation}
R=D_q-D_\nu=\frac{\tau^2}{1-\tau}\,\nabla\log\frac{q_\tau}{\nu_\tau},
\end{equation}
so $\nu=q$ gives $R\equiv0$.
Conversely, let $R=0$ on a nonempty open set $U$ at a level $\tau\in(0,1)$.
The function
\begin{equation}
\ell=\log q_\tau-\log\nu_\tau
\end{equation}
is real-analytic on $\mathbb R^n$ because both densities are positive and real-analytic.
Each partial derivative of $\ell$ vanishes on $U$.
By the identity theorem on the connected set $\mathbb R^n$, these derivatives vanish everywhere.
Thus $\ell$ is constant, and normalization gives $q_\tau=\nu_\tau$.
In characteristic functions,
\begin{equation}
\hat q_\tau(\omega)=\hat q\big((1-\tau)\,\omega\big)\,e^{-\tau^2\|\omega\|^2/2},
\end{equation}
and likewise for $\nu$.
The Gaussian factor never vanishes and $1-\tau>0$, so $\hat q=\hat\nu$ and $q=\nu$.
\hfill$\square$

\noindent The boundary levels are excluded: at $\tau=0$ the observation reveals $X$, while at $\tau=1$ the denoiser returns only $\E_r[X]$.

\noindent\textbf{References that are not clean distributions.}
A stage reference fits the fake model to the forward noising of an intermediate state law $r$ at level $s\le\tau$,
\begin{equation}
\kappa_\tau=P_{s\to\tau}\,r,
\end{equation}
which need not be the noised marginal of any clean distribution.
The clean-prediction form of the stage score is
\begin{equation}
H_\tau[\kappa](\y)=\frac{\y+\tau^2\,\nabla\log\kappa_\tau(\y)}{1-\tau},
\end{equation}
and the proof above gives
\begin{equation}
R_\kappa=D_q-H_\tau[\kappa]=\frac{\tau^2}{1-\tau}\,\nabla\log\frac{q_\tau}{\kappa_\tau},
\end{equation}
which, for $\tau>s$, vanishes on a nonempty open set if and only if $\kappa_\tau=q_\tau$.
For $\tau>s$, the kernel scales by $a>0$ and adds nondegenerate Gaussian noise, so $\kappa_\tau$ is positive and real-analytic.
Characteristic functions give
\[
P_{s\to\tau}r=P_{s\to\tau}r'\quad\Longrightarrow\quad r=r'.
\]

Even with non-uniform fixed weights across steps, assigning positive weight to any earlier step breaks exactness at $q=p$.

\begin{lemma}[Uniqueness of the endpoint reference among fixed mixtures]
\label{lem:app-twoway}
Let $A=D_p$, and let the fake model fit
\begin{equation}
\nu_c=\sum_jc_j\,q_j,\qquad c_j\ge0,\quad\sum_jc_j=1,
\end{equation}
with weights that do not depend on the student.
Call the reference exact if, for every student, the correction vanishes at every query of every step if and only if $q=p$.
Then $\nu_c$ is exact if and only if $c=e_{K-1}$ (the $(K{-}1)$-th standard basis vector), that is, $\nu_c=q$.
\end{lemma}
\noindent\emph{Proof.}
If $c=e_{K-1}$, the correction is $D_p-D_q$ at every query.
At any interior level queried by the final step, zero correction implies $q=p$ by \cref{prop:reference}; conversely, $q=p$ makes the field zero everywhere.
Conversely, if $c_j>0$ for some $j<K-1$, consider a student whose step $j$ outputs $q_j\ne p$ while all other steps (including step $K-1$) output $p$.
Then
\begin{equation}
\nu_c=p+c_j\,(q_j-p)\ne p,
\end{equation}
so $D_p-D_{\nu_c}$ does not vanish on any nonempty open set at any interior level (\cref{prop:reference}), and the student receives a nonzero correction at every interior level despite $q=p$.
\hfill$\square$

\subsection{The endpoint reference and local updates}
\label{app:proofs-cor}

\noindent\textbf{Proof of \cref{cor:endpoint}.}
With $\nu=q$, \cref{eq:split} gives
\begin{equation}
C=A-D_q
\end{equation}
at every query of every step.
Let $A=D_{p^\star}$ at the queried levels.
Because the final step queries an open set at an interior level, zero correction implies $D_q=D_{p^\star}$ and hence $q=p^\star$ by \cref{prop:reference}.
Conversely, if $q=p^\star$, then $C=D_{p^\star}-D_q=0$ everywhere at every queried level, so all steps receive zero correction.
\hfill$\square$

\noindent\textbf{Updates as divergence gradients.}
Applying Tweedie's formula (\cref{eq:app-tweedie}) to $p$ and $q$ gives
\begin{equation}
D_p-D_q=\frac{\tau^2}{1-\tau}\,\nabla_\y\log\frac{p_\tau}{q_\tau}.
\end{equation}
Let a step's clean prediction $\xh$ depend on $\theta$.
Hold its input $\z$ and the query noise fixed, and define
\[
\y=(1-\tau)\xh+\tau\eps,\qquad J=\frac{\partial\xh}{\partial\theta}.
\]
For the final step ($\xh=\x$, so $\y\sim q_\tau$), the reparameterization gradient of $\mathrm{KL}(q_\tau\|p_\tau)$ with respect to the final student call has a zero-mean score-function term and satisfies
\begin{equation}
\begin{aligned}
&-\nabla_\theta\mathrm{KL}(q_\tau\|p_\tau)\\
&\quad=(1-\tau)\,\E\Big[J^\top\nabla_\y\log\frac{p_\tau}{q_\tau}(\y)\Big]\\
&\quad=\mathrm{SNR}(\tau)\,\E\big[J^\top(D_p-D_q)(\y)\big],
\end{aligned}
\end{equation}
where $\mathrm{SNR}(\tau)=(1-\tau)^2/\tau^2$. Up to the normalizer $\omega$ and the scalar weight $\mathrm{SNR}(\tau)$, this matches the final-step term in \cref{eq:dtheta}.
With the mixture or stage reference, $q_\tau$ in the log density ratio is replaced by $\nu_\tau$ or the re-noised stage marginal, whose zero-correction sets are given in \cref{prop:zerosets}.
For an intermediate step, the query law is $q_{k,\tau}$.
With the endpoint density held fixed, the corresponding objective is
\[
\E_{\y\sim q_{k,\tau}}\big[\log q_\tau(\y)-\log p_\tau(\y)\big].
\]
Up to the same positive factor, the local update follows its negative gradient, pushing intermediate queries toward regions where $q_\tau$ under-covers $p_\tau$, and vanishing automatically once $q=p$.

\needspace{9\baselineskip}
\noindent\textbf{Local versus back-propagated endpoint gradients.}
Because \ours detaches the rollout and applies the endpoint correction $D_p-D_q$ locally to each intermediate prediction $\xh_k$, we check when this local update aligns with the endpoint KL gradient back-propagated through later steps.
Let $\Psi_k$ map step $k$'s landing state to the final output.
The output and relevant Jacobians satisfy
\begin{equation}
\begin{aligned}
\x&=\Psi_k(\z_{k+1}),\qquad J_k=\frac{\partial\xh_k}{\partial\theta},\\
M_k&=\frac{\partial\x}{\partial\xh_k}\\
&=\Big(1-\frac{\sigma_{k+1}}{\sigma_k}\Big)\frac{\partial\Psi_k}{\partial\z_{k+1}},
\end{aligned}
\end{equation}
so step $k$'s contribution to the negative gradient of $\mathrm{KL}(q_\tau\|p_\tau)$, holding earlier states fixed and back-propagating through later steps, is
\begin{equation}
\begin{aligned}
&\mathrm{SNR}(\tau)\,\E\big[J_k^\top M_k^\top(D_p-D_q)(\y^{\rm end},\tau)\big],\\
&\hspace{2em}\y^{\rm end}=(1-\tau)\x+\tau\eps.
\end{aligned}
\end{equation}
Step $k$'s local update instead queries $(1-\tau)\xh_k+\tau\eps$ and omits $M_k$.
The expressions coincide if the remaining rollout acts as the identity on $\xh_k$, giving both $\x=\xh_k$ and $M_k=I$.
The following proposition compares their directions in a coordinate-wise Gaussian rollout.

\begin{proposition}[Local and endpoint updates in a Gaussian rollout]
\label{prop:app-local}
Let $p=\mathcal N(0,\Lambda)$ with diagonal $\Lambda$, and let $A=D_p$ and $F=D_q$.
Starting from $\z_0\sim\mathcal N(0,I)$, each student step acts coordinate-wise:
\[
\begin{aligned}
\xh_{k,i}&=a_{k,i}\z_{k,i},\\
\z_{k+1,i}&=g_{k,i}\z_{k,i},\\
g_{k,i}&=r_k+(1-r_k)a_{k,i},\qquad r_k=\frac{\sigma_{k+1}}{\sigma_k}.
\end{aligned}
\]
Fix an intermediate step $k$ and coordinate $i$.
Compare its local update at level $\tau$ with the endpoint update at level $\tau'$ back-propagated through later steps, where
\[
\tau,\tau'\in[\sigma_{k+1},1).
\]
Differentiate both with respect to $a_{k,i}$, holding the other step coefficients fixed and omitting the normalizer.
\begin{enumerate}[label=(\roman*),leftmargin=*,itemsep=3pt]
\item For a landing-state query formed via \cref{eq:app-kernel}, the two updates never have opposite signs.
\item For a fresh-noise query,
\[
\y=(1-\tau)\xh_k+\tau\eps,
\]
the updates never have opposite signs if $a_{k,i}>0$.
They have opposite signs if $a_{k,i}g_{k,i}<0$ and neither update vanishes.
\end{enumerate}
\end{proposition}

\begin{proof}
Positive level factors do not affect the sign comparison.
For a scalar Gaussian, the denoiser is linear:
\[
\begin{aligned}
D_{\mathcal N(0,v)}(y,\tau)&=c_\tau(v)y,\\
c_\tau(v)&=\frac{(1-\tau)v}{(1-\tau)^2v+\tau^2}.
\end{aligned}
\]
Let $\lambda_i=\Lambda_{ii}$ and $v_i=\operatorname{Var}(\x_i)$.
Gaussianity gives
\[
\begin{aligned}
(D_p-D_q)_i(\y,\tau)&=\delta_{\tau,i}\y_i,\\
\delta_{\tau,i}&=c_\tau(\lambda_i)-c_\tau(v_i).
\end{aligned}
\]
Since $c_\tau(v)$ increases in $v$, this difference has the sign of $\lambda_i-v_i$ at every interior level.

The remaining rollout is also linear.
With $\psi_{k,i}=\prod_{j>k}g_{j,i}$, its output and derivative are
\[
\begin{aligned}
\x_i&=\psi_{k,i}g_{k,i}\z_{k,i},\\
\frac{\partial\x_i}{\partial a_{k,i}}&=(1-r_k)\psi_{k,i}\z_{k,i}.
\end{aligned}
\]
At the endpoint query,
\[
\y^{\rm end}=(1-\tau')\x+\tau'\eps,
\]
the expected contribution is
\begin{equation}
\begin{aligned}
&\E\Big[\frac{\partial\x_i}{\partial a_{k,i}}\,\delta_{\tau',i}\y_i^{\rm end}\Big]\\
&\quad=(1-r_k)(1-\tau')\,\delta_{\tau',i}\\
&\qquad\times\psi_{k,i}^2g_{k,i}\,\E\z_{k,i}^2.
\end{aligned}
\end{equation}
Since $\partial\xh_{k,i}/\partial a_{k,i}=\z_{k,i}$, independence of the added noise gives the landing-state local term
\[
\E[\z_{k,i}\delta_{\tau,i}\y_i]
=\frac{1-\tau}{1-\sigma_{k+1}}\,\delta_{\tau,i}g_{k,i}\,\E\z_{k,i}^2.
\]
For a fresh-noise query, it gives
\[
\E[\z_{k,i}\delta_{\tau,i}\y_i]
=(1-\tau)\,\delta_{\tau,i}a_{k,i}\,\E\z_{k,i}^2.
\]
The coefficients $\delta_{\tau,i}$ and $\delta_{\tau',i}$ have the same sign at both levels, so the expected endpoint and landing-state terms cannot oppose.
For fresh noise, opposition occurs when the student coefficient $a_{k,i}$ and landing coefficient $g_{k,i}$ have opposite signs and neither term vanishes.
A positive student coefficient implies a positive landing coefficient.
\end{proof}

At interior noise levels ($\sigma_k<1$), a teacher-initialized denoiser satisfies $a_{k,i}>0$, whereas $a_{0,i}=0$ at $\sigma_0=1$ (where the expected local update vanishes).
Although the sign guarantee is coordinate-wise (shared parameters weight coordinates by $\psi_{k,i}^2$), in the fractal-tree toy (\cref{app:toy-train}) intermediate steps' local updates maintain positive mean cosine similarity with the back-propagated endpoint correction throughout training.

\noindent\textbf{What the update constrains.}
Under the endpoint reference, intermediate steps remain unconstrained at equilibrium so long as the final output matches $p$, whereas the stage reference fixes every intermediate marginal (\cref{prop:zerosets}).
Text-to-image students use this freedom: on \flux, intermediate clean predictions reach $3.3\times$ the final output's norm, and constraining them with the mixture reference degrades final quality (App.~\ref{app:curves}).

\subsection{The mixture reference}
\label{app:proofs-mix}

Let $J\sim\pi$ select a step, with $\pi_k>0$, and draw $X\mid J=k\sim q_k$.
Then $\Y_\tau=(1-\tau)X+\tau\eps$ has law
\begin{equation}
\Y_\tau\sim\nu_\tau=\sum_k\pi_k\,q_{k,\tau}.
\end{equation}
Separating the earlier-step mixture from the endpoint gives
\begin{equation}
\begin{aligned}
\bar q&=\frac{1}{1-\pi_{K-1}}\sum_{k<K-1}\pi_k\,q_k,\\
\nu&=\pi_{K-1}q+(1-\pi_{K-1})\bar q.
\end{aligned}
\end{equation}

\noindent\textbf{Proof of \cref{eq:rmix}.}
Conditioning on the source,
\begin{equation}
\begin{aligned}
D_\nu(\y)&=\E\big[\E[X\mid\Y_\tau,J]\mid\Y_\tau=\y\big]\\
&=\sum_kw_k(\y)\,D_{q_k}(\y),\\
w_k(\y)&=P(J=k\mid\Y_\tau=\y)=\frac{\pi_k\,q_{k,\tau}(\y)}{\nu_\tau(\y)}.
\end{aligned}
\end{equation}
By \cref{prop:reference} with $q_k$ in place of $q$,
\begin{equation}
D_{q_k}-D_\nu=\frac{\tau^2}{1-\tau}\,\nabla\log\frac{q_{k,\tau}}{\nu_\tau},
\end{equation}
and
\begin{equation}
\begin{aligned}
\log P(J=k\mid\y)&=\log\pi_k+\log q_{k,\tau}(\y)\\
&\quad-\log\nu_\tau(\y)
\end{aligned}
\end{equation}
has the same gradient.
For $k=K-1$ this is \cref{eq:rmix}.
\hfill$\square$

\noindent\textbf{The three consequences.}
(1)~If $q=p$, then $R_{\rm mix}=D_p-D_\nu$, and $\nu=p$ holds if and only if $\bar q=p$; otherwise $R_{\rm mix}$ does not vanish on any nonempty open set at any interior level.
(2)~If $\nu=p$, then $R_{\rm mix}=D_q-D_p$, and \cref{eq:split} gives, for any $q$,
\begin{equation}
\begin{aligned}
C&=(A-D_p)+(D_p-D_q)+(D_q-D_p)\\
&=A-D_p.
\end{aligned}
\end{equation}
(3)~With $A=D_p$, $C\equiv0$ holds if and only if $\nu=p$,
\begin{equation}
\pi_{K-1}\,q=p-(1-\pi_{K-1})\,\bar q,
\end{equation}
which is the mixture set of \cref{prop:zerosets} solved for $q$.
Since the right-hand side has total mass $\pi_{K-1}$, it defines a valid distribution $q$ if and only if $p-\sum_{k<K-1}\pi_kq_k$ is a nonnegative measure; wherever $\bar q$ exceeds $p/(1-\pi_{K-1})$, no distribution $q$ can achieve zero correction.
For any differentiable $f$-divergence at the clean level, Theorem~1 of AdaGAN~\cite{tolstikhin2017adagan} gives the exact minimizer when adding a component to a fixed mixture:
\[
q^\star=\frac{[\lambda^\star p-(1-\pi_{K-1})\bar q]_+}{\pi_{K-1}},
\]
with a unique $\lambda^\star$ that normalizes $q^\star$; even when exact cancellation is infeasible, $q^\star$ still zeroes out regions over-represented by $\bar q$ and over-allocates mass elsewhere.
For $K=4$ and $\pi_k=\frac14$, $\nu=p$ reduces to the signed-measure identity
\begin{equation}
q-p=-3\,(\bar q-p),
\end{equation}
so that for every integrable statistic $f$,
\begin{equation}
\E_qf-\E_pf=-3\,\big(\E_{\bar q}f-\E_pf\big).
\end{equation}

\noindent\textbf{The residual as a contrast.}
Group the earlier steps using their posterior weights:
\[
\begin{aligned}
\varpi&=w_{K-1}=P(J=K-1\mid\Y_\tau=\y),\\
D_{\bar q}&=\sum_{k<K-1}\frac{w_k}{1-\varpi}D_{q_k}.
\end{aligned}
\]
Then
\begin{equation}
\begin{aligned}
D_\nu&=\varpi\,D_q+(1-\varpi)\,D_{\bar q},\\
R_{\rm mix}&=D_q-D_\nu=(1-\varpi)\,(D_q-D_{\bar q}),
\end{aligned}
\end{equation}
which matches \cref{eq:rmix} via \cref{prop:reference}.
When $q=p$, $R_{\rm mix}=(1-\varpi)(D_p-D_{\bar q})$ extrapolates $D_p$ away from $D_{\bar q}$, superficially resembling autoguidance~\cite{karras2024autoguidance}.
Under $A=D_p$, $D_\nu=D_p$ gives
\begin{equation}
D_q=D_p+\frac{1-\varpi}{\varpi}\,(D_p-D_{\bar q}),
\end{equation}
so whenever the zero-correction set is attainable, $D_q$ takes the form of an extrapolated denoiser, structurally similar to the fixed points of guided objectives~\cite{luo2025tdm,zheng2025ddo,chen2025gft,zhou2025sidlsg}.
Here, however, the weight $(1-\varpi)/\varpi$ varies with the query, and the direction can reverse with the earlier steps' bias (\cref{sec:toy}), degrading text-to-image quality (App.~\ref{app:curves}).

\noindent\textbf{Endpoint mismatch masked by earlier steps.}
Let $r_L,r_R$ be two distinct distributions (\eg, two branches of the fractal tree), and take uniform step selection with
\begin{center}\footnotesize
\begin{tabular}{@{}lcc@{}}
\toprule
 & mass on $r_L$ & mass on $r_R$ \\
\midrule
teacher $p$ & 0.50 & 0.50 \\
each earlier step $q_0,q_1,q_2$ & 0.40 & 0.60 \\
final step $q_3$ & 0.80 & 0.20 \\
mixture $\nu$ & 0.50 & 0.50 \\
\bottomrule
\end{tabular}
\end{center}
By linearity of the forward noising process,
\begin{equation}
\nu=\tfrac34\,\bar q+\tfrac14\,q_3=p\quad\Longrightarrow\quad\nu_\tau=p_\tau,
\end{equation}
at every noise level, so the mixture reference produces zero correction everywhere even though $q_3\ne p$.
Under the endpoint reference, by contrast, the correction $D_p-D_{q_3}$ is nonzero by \cref{prop:reference}.

\noindent\textbf{The direction follows the earlier steps.}
In one dimension, let $p=\mathcal N(0,1)$, let the three earlier steps output $\mathcal N(0,\beta)$, and let the final step output $\mathcal N(0,v)$ under uniform selection.
The mixture matches the teacher's variance when
\begin{equation}
\frac{3\beta+v}{4}=1\quad\Longleftrightarrow\quad v=1+3\,(1-\beta).
\end{equation}
Any feasible variance match thus has $v>1$ when $\beta<1$ and $v<1$ when $\beta>1$; $v=1$ requires $\beta=1$.
Since mixture moments are $\pi$-weighted averages of step moments, $\nu=p$ forces every finite raw moment of $q$ to deviate from $p$ in the opposite direction to $\bar q$ by a factor of $(1-\pi_{K-1})/\pi_{K-1}$.

\subsection{Zero-correction sets and the stage reference}
\label{app:proofs-sets}

\noindent\textbf{Proof of \cref{prop:zerosets}.}
Endpoint: \cref{cor:endpoint} with $p^\star=p$.
Mixture: every step queries $D_\nu$, so the correction vanishes at every query if and only if $D_\nu=D_p$, \ie, $\nu=p$ (\cref{prop:reference}).
Stage: for $k<K-1$, step $k$ queries $\z_{k+1}$ re-noised to $\tau\in[\sigma_{k+1},\sigma_k]$, with optimal fake $H_\tau[\kappa]$ for
\begin{equation}
\kappa_\tau=P_{\sigma_{k+1}\to\tau}\,r_k .
\end{equation}
The stage correction $D_p-H_\tau[\kappa]$ vanishes if and only if $\kappa_\tau=p_\tau$ (\cref{app:proofs-ref}); since
\begin{equation}
p_\tau=P_{\sigma_{k+1}\to\tau}\,p_{\sigma_{k+1}}
\end{equation}
and $P_{\sigma_{k+1}\to\tau}$ is injective for $\tau>\sigma_{k+1}$, this holds if and only if $r_k=p_{\sigma_{k+1}}$.
For $k=K-1$, the landing state is the final output $\x$, giving $q=p$.
\Cref{lem:app-tdm} below and the matched Euler student in \cref{app:toy} satisfy $q=p$ with $r_k\ne p_{\sigma_{k+1}}$, showing that the stage zero-correction set is a proper subset of $\{q=p\}$.
\hfill$\square$

\begin{lemma}[Stage corrections perturb a matched endpoint]
\label{lem:app-tdm}
Take $n=1$, $p=\mathcal N(0,1)$, and $K=2$ on $(1,s,0)$ with
\begin{equation}
S_u=(1-u)^2+u^2
\end{equation}
denoting the teacher's marginal variance at level $u$.
For $a>0$, let step 0 predict, from $\z_0=Z\sim\mathcal N(0,1)$,
\begin{equation}
\begin{aligned}
\xh_0&=\frac{a-s}{1-s}\,Z,\\
\z_1&=s\,Z+(1-s)\,\xh_0=a\,Z,
\end{aligned}
\end{equation}
and let step 1 output $\x=\z_1/a_0$.
At $a=a_0>0$, the final output has distribution $q=p$.
The intermediate and teacher marginals are
\[
r_0=\mathcal N(0,a_0^2),\qquad p_s=\mathcal N(0,S_s).
\]
If $a_0^2\ne S_s$, the derivative of $\KL(\kappa_\tau\|p_\tau)$ at $a_0$ has the sign of $a_0^2-S_s$ for every $\tau\in(s,1)$.
A sufficiently small gradient step in $a$ therefore perturbs the final variance \mbox{$a^2/a_0^2$} from $1$.
\end{lemma}
\noindent\emph{Proof.}
Set
\[
c=\frac{1-\tau}{1-s},\qquad\eta^2=\tau^2-c^2s^2.
\]
The re-noised marginal is
\begin{equation}
\begin{aligned}
\kappa_\tau&=\mathcal N(0,V),\qquad V=c^2a^2+\eta^2,\\
c^2S_s+\eta^2&=S_\tau.
\end{aligned}
\end{equation}
Then
\begin{equation}
\begin{aligned}
&\frac{\mathrm d}{\mathrm da}\KL\big(\mathcal N(0,V)\,\|\,\mathcal N(0,S_\tau)\big)\\
&\qquad=c^2a\,\Big(\frac{1}{S_\tau}-\frac{1}{V}\Big),
\end{aligned}
\end{equation}
which, for $a>0$, vanishes only when $V=S_\tau$ ($a^2=S_s$) and otherwise has the sign of $a^2-S_s$.
\hfill$\square$

\noindent\textbf{TDM's fake.}
\Cref{prop:zerosets} analyzes TDM's stated objective, where the fake model represents each landing state's marginal noised within the stage's interval.
In practice, TDM trains its fake model differently~\cite{luo2025tdm}: it regresses stage $k$'s clean prediction $\xh_k$ from the re-noised landing state $\Y_\tau=P_{\sigma_{k+1}\to\tau}\z_{k+1}$ using the importance weight
\begin{equation}
w=\frac{g_\tau(\Y_\tau\mid\xh_k)}{g_{\sigma_{k+1}\to\tau}(\Y_\tau\mid\z_{k+1})},
\label{eq:app-tdmw}
\end{equation}
where $g_\tau(\cdot\mid\x)$ is the density of $\x$ noised to $\tau$ and $g_{\sigma_{k+1}\to\tau}$ is the transition density of \cref{eq:app-kernel}.
With exact weight $w$, the objective reduces to denoising $\xh_k$ at level $\tau$ with minimizer $D_{q_k}$, so zero correction requires
\begin{equation}
q_k=p\qquad\text{for every }k.
\end{equation}
TDM's code averages the exponent of $w$ over the $n$ dimensions, giving a weight proportional to $w^{1/n}$ at each $\tau$.
For general $\alpha$, the minimizer is
\begin{equation}
F_\alpha(\y,\tau)=\frac{\E[w^\alpha\,\xh_k\mid\Y_\tau=\y]}{\E[w^\alpha\mid\Y_\tau=\y]},
\end{equation}
which interpolates between $D_{q_k}$ ($\alpha=1$) and $\E[\xh_k\mid\Y_\tau]$ ($\alpha=0$).
Following the teacher's probability flow does not ensure zero correction for these objectives.
For a counterexample covering every $\alpha\in[0,1]$, including $\alpha=1/n$, take a standard Gaussian teacher on the grid $(1,0.9,0.75,0.5,0)$.
Choose the teacher's probability-flow states:
\[
\z_k=\sqrt{S_{\sigma_k}}\,Z,\qquad Z\sim\mathcal N(0,1).
\]
The endpoint then has $q=p$.
To land on these states, the Euler predictions must satisfy
\[
\xh_k=c_kZ,\qquad(c_0,c_1,c_2)\approx(0.055,0.22,0.54).
\]
These intermediate predictions differ from $p$.
At every interior level of stage $k$, this gives
\begin{equation}
\begin{aligned}
\E[\xh_k\mid\Y_\tau]&=m_k\,D_p(\Y_\tau,\tau),\\
m_k&=\frac{c_k\sqrt{S_{\sigma_{k+1}}}}{1-\sigma_{k+1}}.
\end{aligned}
\end{equation}
The three coefficients are approximately 0.50, 0.68, and 0.76.
Let $\pi_0$ denote the joint law of $(\xh_k,\Y_\tau)$ and $\pi_1=w\,\pi_0$ the law under direct noising of $\xh_k$.
These are Gaussian laws; let $\Lambda_0$ and $\Lambda_1$ be their precision matrices.
Since $\z_{k+1}$ is a function of $\xh_k$, the weight depends only on the pair $(\xh_k,\Y_\tau)$.
The weighted law and its precision are therefore
\[
\begin{aligned}
w^\alpha\pi_0&\propto\pi_0^{1-\alpha}\pi_1^\alpha,\\
\Lambda_\alpha&=(1-\alpha)\Lambda_0+\alpha\Lambda_1.
\end{aligned}
\]
Its conditional mean is $F_\alpha=\lambda_kD_p$, where
\begin{equation}
\lambda_k(\alpha,\tau)=-\frac{S_\tau}{1-\tau}\,
\frac{(1-\alpha)\Lambda_{0,xy}+\alpha\Lambda_{1,xy}}
     {(1-\alpha)\Lambda_{0,xx}+\alpha\Lambda_{1,xx}}.
\end{equation}
This ratio is monotone in $\alpha$ because its numerator and positive denominator are affine in $\alpha$.
Its endpoint values are
\[
\begin{aligned}
\lambda_k(0,\tau)&=m_k<1,\\
\lambda_k(1,\tau)&=\frac{c_k^2S_\tau}{(1-\tau)^2c_k^2+\tau^2}<1,
\end{aligned}
\]
where the second inequality uses $c_k<1$.
In this example, every intermediate stage receives a correction $(1-\lambda_k)\,D_p\not\equiv0$ for every $\alpha\in[0,1]$, despite matching the teacher's stage marginals.
\Cref{tab:app-tdmfake} summarizes the four fake objectives; neither stage-marginal nor importance-weighted training preserves $q=p$ in the toy (\cref{app:toy-train}).
\begin{table*}[t]
\centering
\footnotesize
\tabcolsep=4pt
\caption{\textbf{Fake objectives for an Euler rollout.} Training target, optimal fake prediction, and extra conditions for zero correction when $q=p$, with $A=D_p$ and exact denoisers.}
\label{tab:app-tdmfake}
\begin{tabular}{@{}L{3.6cm} L{5.6cm} L{2.4cm} L{4.6cm}@{}}
\toprule
Fake & Training target & Optimal prediction & Extra condition \\
\midrule
Stage law (\cref{prop:zerosets}) & score of $\kappa_\tau=P_{\sigma_{k+1}\to\tau}\,r_k$ & $H_\tau[\kappa]$ & $r_k=p_{\sigma_{k+1}}$, $k<K-1$ \\
TDM, exact weight & $\xh_k$ from re-noised state, weight $w$ & $D_{q_k}$ & $q_k=p$ for every $k$ \\
TDM, released code & $\xh_k$ from re-noised state, weight $w^{1/n}$ & $F_{1/n}$ & can fail even along the teacher's flow \\
Endpoint & final output $\x$ & $D_q$ & none \\
\bottomrule
\end{tabular}
\vspace{-4mm}

\end{table*}

\subsection{Teacher integration in a Gaussian}
\label{app:proofs-pt}

To examine what PT changes, we compare the denoiser with the probability-flow map in one Gaussian coordinate.
This tractable setting explains why trajectory integration sharpens teacher supervision and motivates restricting PT to low noise levels (\cref{eq:pt}).
The denoiser is a conditional mean that shrinks variance; the flow map transports the noisy marginal to the clean distribution.

\begin{lemma}[Denoiser and probability flow]
\label{lem:app-pt}
For an unguided teacher $p=\mathcal N(\mu,\lambda)$ with $\lambda>0$, define
\begin{equation}
\begin{aligned}
S_\tau&=(1-\tau)^2\lambda+\tau^2,\\
\rho_\tau&=\frac{(1-\tau)\sqrt\lambda}{\sqrt{S_\tau}},
\end{aligned}
\end{equation}
where $\rho_\tau$ is the correlation of $X$ and $\Y_\tau$.
\begin{enumerate}[label=(\alph*),leftmargin=*,itemsep=3pt,topsep=3pt]
\item The denoiser is
\begin{equation}
\begin{aligned}
D_p(\y)&=\mu+g_\tau\big(\y-(1-\tau)\mu\big),\\
g_\tau&=\frac{(1-\tau)\lambda}{S_\tau}.
\end{aligned}
\end{equation}
\item The exact probability-flow map to level $0$ is
\begin{equation}
\begin{aligned}
\Phi(\y)&=\mu+h_\tau\big(\y-(1-\tau)\mu\big),\\
h_\tau&=\sqrt{\lambda/S_\tau},
\end{aligned}
\end{equation}
with $h_\tau/g_\tau=1/\rho_\tau>1$.
\item For $\Y_\tau\sim p_\tau$, the output variances are
\begin{equation}
\begin{aligned}
\operatorname{Var}D_p(\Y_\tau)&=\rho_\tau^2\lambda,\\
\operatorname{Var}\Phi(\Y_\tau)&=\lambda,
\end{aligned}
\end{equation}
and the reconstruction errors are
\begin{equation}
\begin{aligned}
\E|D_p(\Y_\tau)-X|^2&=\lambda(1-\rho_\tau^2),\\
\E|\Phi(\Y_\tau)-X|^2&=2\lambda(1-\rho_\tau).
\end{aligned}
\end{equation}
The denoiser minimizes reconstruction error, while the flow map preserves the teacher's marginal variance.
\item For $q=\mathcal N(\mu,v)$, the denoiser coefficient is
\begin{equation}
g_\tau(v)=\frac{(1-\tau)\,v}{(1-\tau)^2v+\tau^2}.
\end{equation}
It increases in $v$, with $g_\tau(v)=h_\tau$ if and only if \mbox{$v=\lambda(1+1/\rho_\tau)$}.
\end{enumerate}
\end{lemma}
\begin{proof}
(a)~Apply Gaussian conditioning with
\[
\mathrm{Cov}(X,\Y_\tau)=(1-\tau)\lambda,\qquad
\operatorname{Var}\Y_\tau=S_\tau.
\]

\noindent (b)~Write the flow in centered coordinates:
\[
\frac{\mathrm d\y_t}{\mathrm dt}=\E[\eps-X\mid\Y_t=\y_t],\qquad
u_t=\y_t-(1-t)\mu.
\]
Part (a) gives
\begin{equation}
\begin{aligned}
\frac{\mathrm du_t}{\mathrm dt}
&=\frac{t-(1-t)\lambda}{S_t}\,u_t\\
&=\frac12\,\frac{\mathrm d\log S_t}{\mathrm dt}\,u_t,
\end{aligned}
\end{equation}
hence
\begin{equation}
u_0=\sqrt{S_0/S_\tau}\;u_\tau,\qquad S_0=\lambda.
\end{equation}

\noindent (c)~The variances follow from $g_\tau^2S_\tau$ and $h_\tau^2S_\tau$.
The errors use $h_\tau(1-\tau)\lambda=\rho_\tau\lambda$; their difference is $\lambda(1-\rho_\tau)^2\ge0$.

\noindent (d)~Direct differentiation gives
\begin{equation}
\partial_v\,g_\tau(v)=\frac{(1-\tau)\,\tau^2}{\big((1-\tau)^2v+\tau^2\big)^2}>0.
\end{equation}
To solve $g_\tau(v)=h_\tau$, use the identities
\[
(1-\tau)h_\tau=\rho_\tau,\qquad
\frac{\tau^2}{(1-\tau)^2\lambda}=\frac{1-\rho_\tau^2}{\rho_\tau^2}.
\]
They give
\[
\frac{v}{\lambda}=\frac{1+\rho_\tau}{\rho_\tau}.\qedhere
\]
\end{proof}

\noindent\textbf{A single noise level.}
At $\lambda=1$ and $\tau=\frac12$, $\rho_\tau=1/\sqrt2$, so
\begin{equation}
\begin{aligned}
g_{1/2}&=1,\qquad h_{1/2}=\sqrt2,\\
v^\star&=(1+\sqrt2)\lambda.
\end{aligned}
\end{equation}
Here $v^\star$ matches the flow coefficient at this level, as in part (d); the denoiser and flow-map reconstruction errors are $0.5$ and $2-\sqrt2\approx0.586$, respectively.
The ratio $v^\star/\lambda=1+1/\rho_\tau$ increases with $\tau$ and diverges as $\tau\to 1$; for $\lambda=1$, its values at $\tau=0.25,0.5,0.9$ are $2.05,2.41,10.06$, explaining why applying PT at high noise destabilizes training (App.~\ref{app:impl}).

\noindent\textbf{Finite-step integration.}
PT uses four Euler steps on $t_j=\tau(1-j/4)$, $j=0,\dots,4$; the last step returns the denoiser prediction at $\tau/4$.
For $p=\mathcal N(0,\lambda)$, its coefficient is
\begin{equation}
c_4=\prod_{j=0}^{3}\bigg[1+(t_{j+1}-t_j)\frac{t_j-(1-t_j)\lambda}{S_{t_j}}\bigg].
\end{equation}
It lies between $g_\tau(\lambda)$ and $h_\tau$.
Introduce
\[
\mathbf u_t=((1-t)\sqrt\lambda,t),\qquad \|\mathbf u_t\|^2=S_t.
\]
Each Euler factor is
\[
\frac{\mathbf u_{t_{j+1}}\!\cdot\mathbf u_{t_j}}{\|\mathbf u_{t_j}\|^2}
=\frac{\|\mathbf u_{t_{j+1}}\|}{\|\mathbf u_{t_j}\|}\cos\theta_j.
\]
Multiplying them gives
\begin{equation}
\begin{aligned}
c_4&=\frac{\|\mathbf u_0\|}{\|\mathbf u_\tau\|}\prod_{j=0}^{3}\cos\theta_j\\
&=h_\tau\prod_{j=0}^{3}\cos\theta_j.
\end{aligned}
\end{equation}
Here $\theta_j>0$ is the angle between successive vectors, with
\[
\sum_j\theta_j=\arccos\rho_\tau<\pi/2.
\]
For positive angles with sum below $\pi/2$, the cosine addition formula gives
\[
\rho_\tau<\prod_j\cos\theta_j<1.
\]
Hence
\[
g_\tau(\lambda)=\rho_\tau h_\tau<c_4<h_\tau.
\]
The same argument applies to any strictly decreasing Euler grid with at least two steps.
At $\lambda=1$ and $\tau=\frac12$, $c_4\approx1.30$ (matching variance $1.88\lambda$) lies between $g_{1/2}=1$ and $h_{1/2}\approx1.41$ ($2.41\lambda$).

The preceding analysis treats one noise level at a time.
The next lemma treats the low-noise integration and higher-noise denoiser targets together, as in \cref{eq:pt}.

\begin{lemma}[Bounds on the equilibrium variance]
\label{lem:app-balance}
For $p=\mathcal N(0,\lambda)$, consider a scale-only final step with an exact endpoint fake:
\[
\x=\sqrt v\,Z,\qquad Z\sim\mathcal N(0,1),\qquad q=\mathcal N(0,v).
\]
Take a compact query interval $\mathcal T\subset(0,1)$ containing $s=\sigma_{K-1}$ in its interior.
The teacher coefficient is
\[
\tilde c_\tau=\begin{cases}
c_\tau,&\tau<s,\\
g_\tau(\lambda),&\tau\ge s,
\end{cases}
\]
where $c_\tau$ is measurable and $g_\tau(\lambda)<c_\tau\le h_\tau$.
For positive level weights $w_\tau(v)$ jointly continuous in $(\tau,v)$, the expected update of $\sqrt v$ is proportional to
\begin{equation}
U(v)=\int_{\mathcal T}w_\tau(v)\,(1-\tau)\,\big(\tilde c_\tau-g_\tau(v)\big)\,\mathrm d\tau.
\end{equation}
This function has a root satisfying
\[
\lambda<v<(1+1/\rho_s)\lambda.
\]
The root is unique if $w_\tau$ is independent of $v$.
\end{lemma}
\noindent\emph{Proof.}
For a fresh-noise endpoint query, the correction and expected update are
\begin{equation}
\begin{aligned}
\Y_\tau&=(1-\tau)\x+\tau\eps,\\
C&=\big(\tilde c_\tau-g_\tau(v)\big)\Y_\tau,\\
\E\Big[C\,\frac{\partial\x}{\partial\sqrt v}\Big]
&=\big(\tilde c_\tau-g_\tau(v)\big)(1-\tau)\sqrt v.
\end{aligned}
\end{equation}
This gives $U(v)$ up to the positive factor $\sqrt v$.
At $v=\lambda$, the integrand is nonnegative and is strictly positive below $s$, so $U(\lambda)>0$.
At $v_{\max}=(1+1/\rho_s)\lambda$, monotonicity of $\rho_\tau$ and part (d) give, for $\tau<s$,
\begin{equation}
\begin{aligned}
g_\tau(v_{\max})&\ge g_\tau\big((1+1/\rho_\tau)\lambda\big)\\
&=h_\tau\ge c_\tau.
\end{aligned}
\end{equation}
Above $s$, $g_\tau(v_{\max})>g_\tau(\lambda)$, so $U(v_{\max})<0$.
Continuity of $U$ yields a root between $\lambda$ and $v_{\max}$.
When $w_\tau$ is independent of $v$, strict increase of $g_\tau(v)$ makes $U$ strictly decreasing, giving uniqueness.
\hfill$\square$

\noindent With $\lambda=1$, $\tau\sim\mathcal U[0.02,0.99)$, uniform weights, and $s=0.5$, restricting PT to $\tau<s$ gives this scale-only student an equilibrium variance of approximately $1.38\lambda$ under exact integration and $1.27\lambda$ under four Euler steps, well below the upper bound $2.41\lambda$.
These values depend on the teacher targets and query weights, rather than the student's earlier steps.

\noindent\textbf{Teacher enhancement and residual cancellation.}
The endpoint reference removes the reference residual for any teacher target, leaving $C=A-D_q$ at every query.
The further characterization $q=p^\star$ in \cref{cor:endpoint} applies when $A=D_{p^\star}$ for one clean distribution across queried levels.
Guided and trajectory-integrated targets generally specify a different family of predictions.
For CFG, let $p^u$ be the unconditional distribution.
At each level, the guided target has the Tweedie form for
\[
\pi_\tau\propto p_\tau^{\,w}(p^u_\tau)^{1-w},
\]
whenever normalizable.
This family generally is not the forward noising of one clean distribution~\cite{bradley2024cfgpc,zheng2024characteristic}.
Likewise, the unguided Gaussian flow map at $\tau$ equals the denoiser of
\[
\mathcal N\big(\mu,(1+1/\rho_\tau)\lambda\big),
\]
whose variance depends on $\tau$; \cref{eq:pt} also switches targets at $\sigma_{K-1}$.
These enhancements change the teacher supervision while preserving the cancellation $C=A-D_q$; for the Gaussian example, \cref{lem:app-balance} characterizes the resulting balance across levels.
The Gaussian analysis above considers $w=1$; in practice, guided PT integrates the guided field $B$ (\cref{sec:teacher}).

\section{Toy construction}
\label{app:toy}

We detail the teacher and student models (\cref{app:toy-students}), single-step updates and directional reversal (\cref{app:toy-update}), teacher targets on the final interval (\cref{app:toy-answer}), and full training under each reference (\cref{app:toy-train}) for \cref{fig:toy}.

\subsection{Teacher and students}
\label{app:toy-students}

\begin{figure*}[t]
\centering
\includegraphics[width=.95\textwidth]{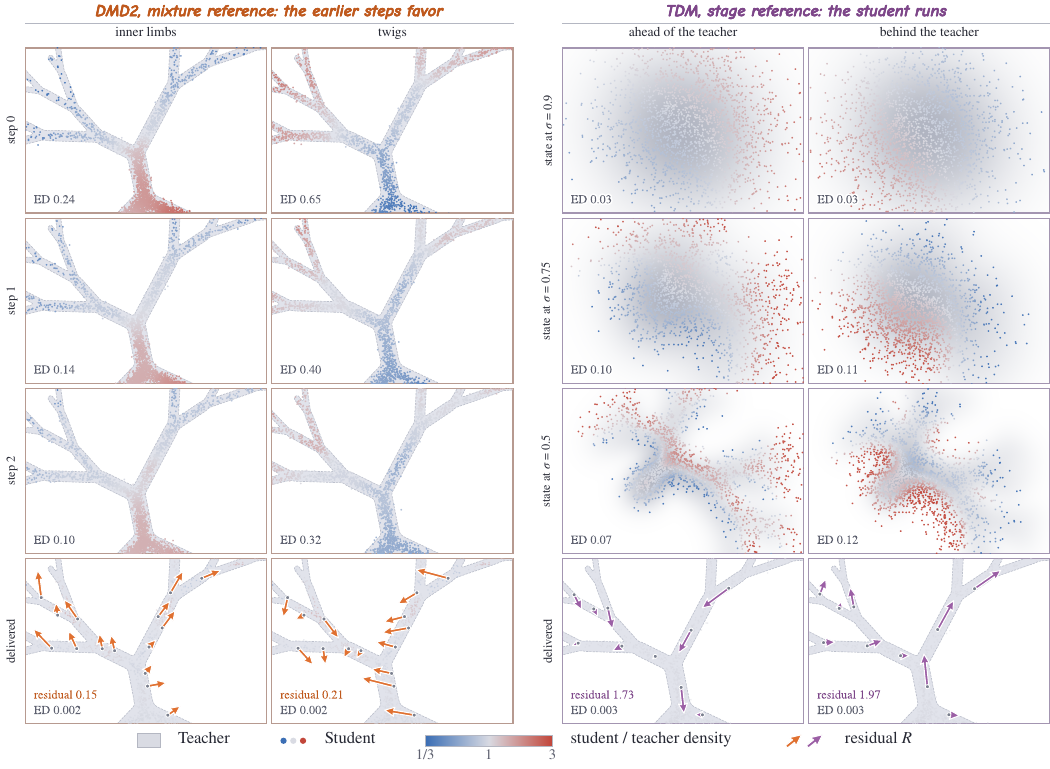}
\caption{\textbf{Opposite updates despite correct final outputs.}
Each pair has $q=p$ but opposite intermediate biases, shown in the top three rows: clean predictions (DMD2, left) or landing states (TDM, right).
Gray backgrounds show the teacher; dots are student samples, colored by relative density (blue: below the teacher; red: above).
Bottom: updates induced by the reference residual (numbers: mean norms); the endpoint reference gives zero update.
ED: energy distance (lower is better).}
\label{fig:app-toy-students}
\end{figure*}

\noindent\textbf{Teacher and grid.}
The teacher is a rescaled version of class A in the fractal-tree benchmark of~\citet{karras2024autoguidance} with one fewer subdivision, comprising 63 branches of 8 two-dimensional Gaussian components each (504 in total).
Its noised densities and denoisers are available in closed form; probability-flow trajectories are computed by numerical integration.
All students use $A=D_p$ and the \sdthree grid $(1,0.9,0.75,0.5,0)$.
\par\needspace{8\baselineskip}\noindent
We evaluate a distribution $q$ by its energy distance to $p$,
\begin{equation}
\begin{aligned}
\mathrm{ED}(q,p)&=2\,\E\|X-Y\|\\
&\quad-\E\|X-X'\|-\E\|Y-Y'\|,
\end{aligned}
\end{equation}
with independent $X,X'\sim q$ and $Y,Y'\sim p$, averaged over 8 sets of 4{,}096 student samples against 32{,}768 teacher samples (sampling floor 0.002--0.003).

\noindent\textbf{Consistency students.}
The DMD2 row uses consistency students (\cref{fig:app-toy-students}, left) with frozen intermediate steps 0--2.
Each step $k$ integrates from its input to the endpoint of a tilted teacher ODE with component weights
\begin{equation}
w_j^{(k)}\propto w_j\,e^{\beta_ku_j},
\end{equation}
where $w_j$ is component $j$'s weight in $p$ and $u_j$ is its standardized distance from the tree center.
Setting $\beta=-(0.6,0.4,0.2)$ biases steps 0--2 toward the inner limbs (ED 0.24, 0.14, 0.10), whereas $\beta=(0.6,0.4,0.2)$ biases them toward the outer twigs (ED 0.65, 0.40, 0.32).
The ODE is integrated using 48, 32, and 32 Heun substeps for steps 0, 1, and 2 (at $\beta=0$, every step's output lies at the ED sampling floor).
\Cref{fig:app-toy-students} colors each step's outputs by the log density ratio of their kernel density estimate (bandwidth 0.45) to the identically smoothed teacher density.
In the right column of \cref{fig:toy}, step 3 outputs exact teacher samples ($q=p$); in the middle column, it outputs the midpoint between the teacher's posterior mean and ODE endpoint from $\z_3$, producing a blurred distribution that inherits the earlier steps' inward bias (ED 0.11).

\noindent\textbf{Euler students.}
The TDM and \ours rows use Euler students (\cref{fig:app-toy-students}, right).
The middle column of \cref{fig:toy} uses the teacher-initialized Euler student, where each step predicts the teacher's posterior mean $D_p(\z_k,\sigma_k)$ (ED 0.15).
The right column uses a matched Euler student with faster intermediate denoising (labeled ``ahead of the teacher'' in \cref{fig:app-toy-students}), whose intermediate state at $\sigma_k$ is the teacher's exact ODE state at noise level $a_k$,
\begin{equation}
\z_k=\Phi_{1\to a_k}(\z_0),
\end{equation}
where $(a_1,a_2,a_3)=(0.85,0.65,0.40)$ and $\Phi$ is the teacher's probability-flow map.
Its final output is an exact teacher ODE sample (ED 0.003, at the sampling floor), while its intermediate marginals have ED 0.03, 0.10, and 0.07 to $p_\sigma$ at $\sigma=0.9$, $0.75$, and $0.5$.
Its slower-denoising counterpart (``behind the teacher'' in \cref{fig:app-toy-students}) uses $(a_1,a_2,a_3)=(0.95,0.85,0.62)$, with intermediate EDs 0.03, 0.11, and 0.12.
\Cref{fig:app-toy-students} colors each landing state by the exact log density ratio $\log(p_{a_k}/p_{\sigma_k})$.

\subsection{One update}
\label{app:toy-update}

\noindent\textbf{References and single-step updates.}
In the DMD2 row, the mixture fake averages four equal-weight components: kernel density estimates (bandwidth 0.05) from 6{,}000 rollouts of each intermediate step and the endpoint fake.
For the blurred student, the endpoint fake is a kernel density estimate of final outputs; for the matched student ($q=p$), both the endpoint fake and teacher target use $p$ convolved with the same kernel, so $A-D_q\equiv 0$.
Each final sample's update is the mean of $A-F$ over 24 fresh-noise queries on $\tau\sim\mathcal U[0.02,0.99)$, decomposed into the endpoint correction $A-D_q$ (gray arrow) and the mixture residual $R$ (colored arrow).
In the Euler rows, each step's correction is averaged over 24 queries per rollout: TDM re-noises $\z_{k+1}$ via $P_{\sigma_{k+1}\to\tau}$ to $\tau\sim\mathcal U(\sigma_{k+1},\sigma_k]$, whereas \ours queries $\xh_k$ with fresh noise on $[\sigma_{k+1},0.99)$ (or $[0.02,0.99)$ for $k=3$).
We perturb each prediction by $\eta=0.1$ times its mean correction, recompute subsequent steps, and plot the resulting displacement of $\x$ per unit $\eta$ under the endpoint reference (gray arrow) and the displacement difference between TDM's stage reference and the endpoint reference (colored arrow).
For the teacher-initialized student, references are kernel densities from 32{,}768 rollouts (bandwidth 0.02); for the faster-denoising matched student, all references are exact closed-form Gaussian mixtures, so the endpoint correction vanishes identically.
As a finite-sample sanity check, applying the same kernel-density estimators to an untilted consistency student or an exact ODE Euler student yields small estimation floors (residual norm 0.013 vs.\ 0.15--0.21 in DMD2; desired and residual norms 0.05 and 0.12 vs.\ 0.45 and 0.48 for the teacher-initialized Euler student).

\noindent\textbf{Numbers in \cref{fig:toy}.}
The numbers in each panel report the mean vector norm over 2{,}000 final samples (unnormalized correction in the DMD2 row; displacement per unit $\eta$ in the Euler rows) and should be compared within each row.

\noindent\textbf{Direction.}
Evaluating the single-step update on both pairs of matched students ($q=p$, bottom row of \cref{fig:app-toy-students}) at shared final samples ($N=2{,}000$) shows that the updates induced by the reference residual reverse with the intermediate bias:
across three seeds, the two DMD2 students (inner limbs vs.\ outer twigs) receive updates with mean norms 0.15 and 0.21, median pointwise cosines from $-0.91$ to $-0.97$, and negative pointwise cosines at 85--94\% of samples.
For TDM (faster- vs.\ slower-denoising matched students), the displacements have mean norms 1.73 and 1.97, median cosine $-0.999$, and negative pointwise cosines at 96.6--97.2\% of samples across three seeds, whereas the endpoint reference yields zero update for both pairs.

\subsection{Teacher targets on the final interval}
\label{app:toy-answer}

Using the fork region in the bottom row of \cref{fig:overview}, we compare four noise levels below $\sigma_3=0.5$: $\tau\in\{0.15,0.25,0.32,0.45\}$, with shared clean samples and noise (2{,}000 queries per level).
A target is on-tree if its log density under $p$ exceeds the first percentile on 20{,}000 teacher samples.
Single-step predictions $D_p(\y,\tau)$ are on-tree 96\%, 83\%, 66\%, and 42\% of the time, compared with 99\%, 98\%, 97\%, and 96\% for PT.
Here a single denoiser evaluation averages across unresolved branches, while trajectory integration retains their fine structure in the final-step targets.

\subsection{Training under each reference}
\label{app:toy-train}

\begin{figure*}[t]
\centering
\includegraphics[width=.95\textwidth]{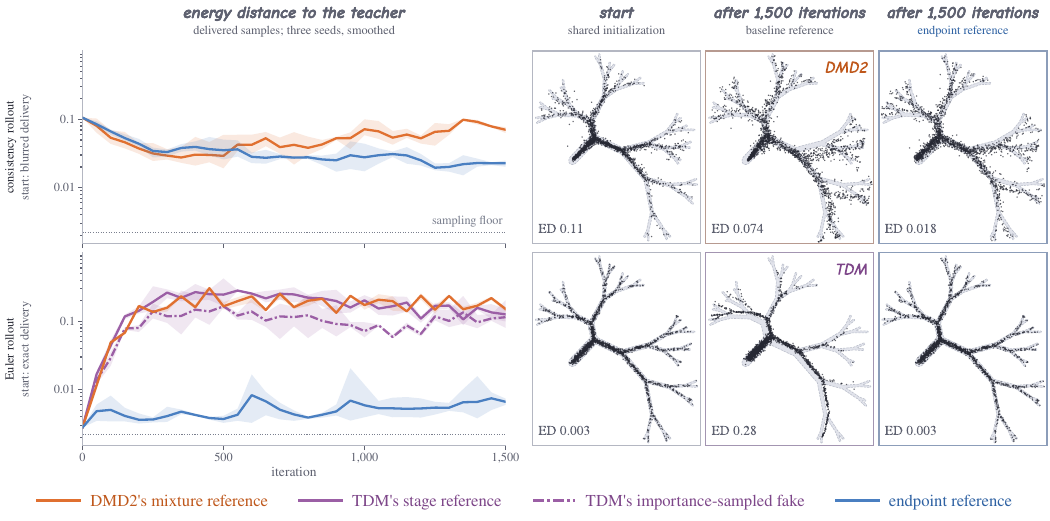}
\caption{\textbf{The endpoint reference improves imperfect students and preserves matched outputs.}
Top: a consistency student starts with blurred outputs; bottom: an Euler student starts at $q=p$.
Left: final output energy distance (lower is better; smoothed geometric mean and range over three seeds).
Right: seed-0 samples before and after 1{,}500 iterations with the baseline or endpoint reference (shown with single-snapshot ED at iteration 1{,}500). Gray: teacher; black dots: student.}
\label{fig:app-toy-train}
\end{figure*}

\noindent\textbf{Setup.}
As in text-to-image distillation, all four steps share a residual network with a zero-initialized output layer, conditioned on the step's noise level.
We train all steps for 1{,}500 iterations across three seeds using Adam ($\beta=(0,0.99)$, learning rate $3\times10^{-4}$, gradient clip 3, batch size 512), unguided teacher target $D_p$, a per-step mini-batch normalizer $\omega$ in \cref{eq:update}, and kernel-density fake models over the last 8{,}192 samples (bandwidth 0.03), querying the full noise range for consistency rollouts (as in DMD2) and above $\sigma_{k+1}$ for Euler rollouts (as in \ours).
We report median $\mathrm{ED}(q,p)$ over iterations 500--1{,}500 across seeds (\cref{fig:app-toy-train} evaluates every 50 iterations with EMA smoothing 0.35; floor 0.0022).

\noindent\textbf{From an imperfect initialization.}
In the consistency rollout, DMD2's mixture reference reduces the 4-step mixture distance from 0.14 to 0.03--0.04 but leaves the final output at $\mathrm{ED}(q,p)=0.063$, whereas the endpoint reference reaches 0.027 and restores the share of samples on the outermost two tree levels from 11\% to 19\% (teacher: 20\%).
In the Euler rollout from the teacher-initialized student (not shown), TDM's stage reference settles at 0.045 and the mixture reference at 0.33, whereas the endpoint reference reaches 0.021.

\noindent\textbf{From a matched student ($q=p$).}
Initialized at the faster-denoising matched student ($\mathrm{ED}(q,p)$ at the sampling floor), TDM's stage-marginal fake, TDM's exact-importance fake (``importance-sampled'' in \cref{fig:app-toy-train}), and the mixture fake all drive the final output away from $p$ (median $\mathrm{ED}(q,p)$ of 0.18, 0.092, and 0.19).
For the stage reference, this happens while intermediate states move closer to the teacher's marginals (their mean intermediate ED across the three stages falls from 0.066 to 0.018--0.028).
The endpoint reference keeps $\mathrm{ED}(q,p)$ at 0.004.

\noindent\textbf{Local updates of intermediate steps.}
Comparing the intermediate steps' local updates in \cref{eq:update} against the final-step correction back-propagated through the rollout on the same batch (excluding step 3's local term from both), their mean cosine similarity across three seeds stays positive throughout training (empirically confirming the alignment analyzed in \cref{prop:app-local}): 0.22--0.73 when starting from the teacher-initialized student and 0.83--0.98 when starting from the faster-denoising matched student.

\section{Implementation details}
\label{app:impl}

\Cref{tab:app-hyper} summarizes the shared hyperparameters and teacher-specific guidance scales, grids, learning rates, and schedules (\cref{alg:etdm,sec:impl}); App.~\ref{app:qwen} gives the remaining settings for \qwen.
Fake-training targets and intermediate rollout states are detached, so student gradients flow only through the selected step's clean prediction.

\noindent\textbf{Euler and DDIM steps.}\phantomsection\label{app:euler-ddim}
Under linear noising, a prediction can be parameterized as a clean sample, noise, or velocity.
At state $\z_k$, these parameterizations satisfy
\[
\begin{aligned}
\z_k&=(1-\sigma_k)\xh_k+\sigma_k\hat\eps_k,\\
\hat{\mathbf v}_k&=\hat\eps_k-\xh_k=\frac{\z_k-\xh_k}{\sigma_k}.
\end{aligned}
\]
Substituting into the deterministic DDIM transition gives
\[
\begin{aligned}
\z_{k+1}&=(1-\sigma_{k+1})\xh_k+\sigma_{k+1}\hat\eps_k\\
&=\frac{\sigma_{k+1}}{\sigma_k}\z_k
+\Big(1-\frac{\sigma_{k+1}}{\sigma_k}\Big)\xh_k,
\end{aligned}
\]
which is the Euler step in \cref{eq:euler}.
TDM refers to this deterministic rollout as DDIM~\cite{luo2025tdm}, whereas DMD2 and D-DMD draw fresh noise for $\hat\eps_k$ at each step.

\noindent\textbf{Fake capacity.}
On \fluxnine, the main \ours run uses a full-parameter fake (\cref{tab:app-hyper}), whereas locally trained baselines use rank-32 LoRA fakes (App.~\ref{app:baselines}).
Controlling for fake capacity with a rank-32 LoRA fake, \ours still reaches 12.00 \hpspp and exceeds every ranked baseline in \cref{tab:main} on all metrics except Aesthetic.

\begin{table}[t]
\centering
\footnotesize
\tabcolsep=4pt
\caption{\textbf{Training settings.} $\mathrm{SNR}(\tau)=((1-\tau)/\tau)^2$. FLUX: both FLUX.2-klein models.}
\label{tab:app-hyper}
\begin{tabular}{@{}L{1.95cm} L{5.95cm}@{}}
\toprule
Item & Value \\
\midrule
Student & LoRA~\cite{hu2022lora} on the teacher, rank 32, $\alpha=64$ \\
Fake & as the student; \fluxnine: all parameters \\
Student optimizer & AdamW~\cite{loshchilov2019adamw}, lr $6\times10^{-6}$ (\fluxnine: $3\times10^{-6}$), $\beta=(0,0.95)$, weight decay 0.01, clip 3 \\
Student schedule & 100 warm-up updates, then constant; \sdthree and \fluxnine: cosine decay to 0 from update 1{,}200 \\
Fake optimizer & AdamW, lr $2\times10^{-5}$ (\fluxnine: $1.5\times10^{-6}$), $\beta=(0,0.95)$, weight decay 0.01, clip 1, constant \\
Fake updates & 5 per student update, each on fresh prompts and a fresh rollout; the student reuses the last rollout \\
Fake loss & \cref{eq:critic}, $\tau\sim\mathcal{U}[0.02,0.99)$, weight $\min\{\max\{\mathrm{SNR}(\tau),1\},10\}$ \\
Rollout & deterministic Euler on the native grid \\
Step choice & $k\sim\mathcal{U}\{0,\dots,K{-}1\}$ per sample \\
Steps $k<K{-}1$ & $\tau\sim\rho_k$ on $[\sigma_{k+1},0.99)$ (\cref{eq:law}); $\y=(1-\tau)\xh_k+\tau\eps$, $\eps$ fresh \\
Step $K{-}1$ & $\tau\sim\mathcal{U}[0.02,0.99)$; $\y=(1-\tau)\x+\tau\eps$ \\
Teacher target & $\Tw$; for $k=K{-}1$ and $\tau<\sigma_{K-1}$, PT (\cref{eq:pt}): four guided Euler steps from $\tau\{1,\tfrac34,\tfrac12,\tfrac14\}$ \\
Normalizer & $\omega=\operatorname{mean}|\xh_k-A|$ per sample (\cref{eq:update})~\cite{yin2024dmd} \\
Guidance & teacher default: \sdthree $w=7$, FLUX $w=4$ \\
Grid & \sdthree $\sigma=1,0.9,0.75,0.5,0$; FLUX $1,0.967,0.908,0.767,0$ \\
Batch, resolution & 16 prompts, $1024^2$ \\
Updates, EMA & 2{,}000; EMA decay 0.999 \\
\bottomrule
\end{tabular}

\end{table}

\noindent\textbf{Cost and scope of PT.}
PT applies to final-step queries below $\sigma_{K-1}$.
It integrates the guided field on \sdthree and FLUX; on \qwen it integrates $\Tc$, since PAG is confined to higher levels (App.~\ref{app:qwen}).
Under uniform step selection and $\tau\sim\mathcal{U}[0.02,0.99)$ for step $K-1$, the fraction of queries falling in this interval is
\begin{equation}
f_{\rm PT}=\frac{\sigma_{K-1}-0.02}{0.97\,K},
\end{equation}
or 12.4\% on \sdthree and 19.3\% on FLUX.
Selective evaluation would add $3f_{\rm PT}$ extra guided teacher calls per sample in expectation.
Our reported timings use a simple masked implementation that evaluates PT on the full query batch rather than selectively.
The final-step, low-noise restriction matters: extending PT to all queries of every step causes training to collapse on \flux (reaching an absolute \hpspp of $-1.23$ at 1k updates) and lowers \hpspp by 1.43 on \sdthree at 500 updates.

\noindent\textbf{What PT changes.}
PT modifies both the teacher target $A$ and the normalizer $\omega=\operatorname{mean}|\xh_k-A|$.
To separate trajectory integration from normalizer rescaling and multi-step CFG, we run two control experiments.
In shorter runs, using only PT's normalizer leaves \hpspp unchanged relative to w/o PT on \sdthree (300--900 updates) and 0.07 lower on \flux (300--600).
Integrating only the conditional field $\Tc$ while keeping a single-step CFG term matches the full method on \sdthree (300--2k) and recovers 0.46 of its 0.53 mean gain on \flux (300--1.3k).
These controls support trajectory integration as the main source of PT's gain.

\noindent\textbf{Back-propagating through the rollout.}
To test whether detaching the rollout omits useful multi-step gradient signal, we evaluate back-propagating the endpoint correction through the rollout.
Adding the back-propagated endpoint correction to local updates lowers \hpspp by 0.28 on \sdthree and 1.74 on \flux on average over matched checkpoints from 300 updates onward.
In these runs, intermediate steps account for 75--88\% of the summed back-propagated velocity gradient norms and trigger more frequent gradient clipping.
Using only back-propagated endpoint supervision reaches peak scores of 5.01 on \flux and 13.60 on \qwen, compared with 12.70 and 14.69 for \ours.

\noindent\textbf{Compute and training time.}
On 16 NPUs (batch 16 at $1024^2$), median per-update times for \ours (\cref{alg:etdm}) are 8.3\,s on \sdthree, 17.6\,s on \flux, and 53.6\,s on \fluxnine (4.6, 9.8, and 29.8 hours for 2{,}000 updates, excluding evaluation).
Query calibration (App.~\ref{app:calib}) takes 5.4, 15.8, and 33.7 minutes.

\begin{figure*}[t]
\centering
\includegraphics[width=\textwidth]{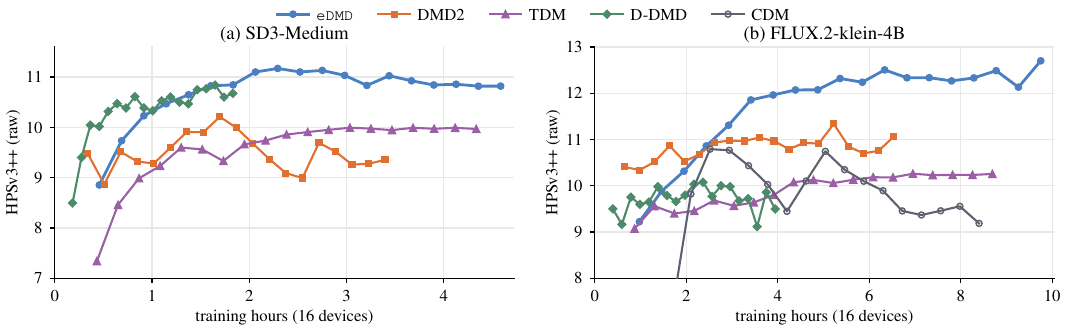}
\caption{\textbf{\hpspp versus wall-clock training time.}
Raw weights evaluated every 100 updates from 200 onward; time equals updates $\times$ median per-update time on 16 devices
(\sdthree: \ours 8.3\,s, DMD2 6.1\,s, D-DMD 3.3\,s, TDM 7.8\,s; \flux: 17.6, 11.8, 7.1, 15.6, and CDM 15.1\,s).
All runs terminate at 2{,}000 updates. On \sdthree, TDM is our re-implementation; the official CDM checkpoint has no training curve.}
\label{fig:app-time}
\end{figure*}

In wall-clock time (\cref{fig:app-time}), \ours on \sdthree surpasses the peak of our TDM re-implementation within 0.9 hours and that of DMD2 within 1.1 hours, and matches D-DMD's peak (10.84) in about 1.8 hours (the time D-DMD requires for 2{,}000 updates).
On \flux, \ours surpasses every baseline's peak within 3.4 hours, before DMD2 reaches its own peak at 5.2 hours.

\noindent\textbf{Evaluation.}
We render DrawBench-1000 case $i$ with seed $42+i$ at $1024^2$ in bf16 using the teacher's VAE.
CLIP score is the unscaled cosine similarity between OpenAI CLIP ViT-B/32 image and text embeddings~\cite{radford2021clip}.
We generate four $1024^2$ images per DPG-Bench~\cite{hu2024ella} prompt and score them with the official mPLUG-large QA script.

\section{Calibration of the query density}
\label{app:calib}

Before training, we calibrate the intermediate query density (\cref{eq:law}) once per teacher and grid using teacher samples alone.
To compare predictions across levels, let $X$ be a clean sample and $\Y_s=(1-s)X+s\eps$, with $\eps\sim\mathcal N(0,I_n)$.
For $s\le\tau<1$, further noising gives
\begin{equation}
\tilde\Y_\tau=a\,\Y_s+\sqrt{\tau^2-a^2s^2}\,\bm\zeta,
\label{eq:forward}
\end{equation}
where $a=(1-\tau)/(1-s)$ and $\bm\zeta\sim\mathcal N(0,I_n)$ is independent of $X$ and $\eps$.
Conditional on $X$, $\tilde\Y_\tau$ has the same law as a fresh-noise query $(1-\tau)X+\tau\eps$.
This coupling applies to noised teacher samples rather than the student's Euler states.
For an exact unguided teacher with $X\sim p$, conditioning gives
\[
D_p(\tilde\Y_\tau,\tau)=\E[D_p(\Y_s,s)\mid\tilde\Y_\tau].
\]
Thus consistent predictions may differ through averaging; our score estimates drift beyond this expected change.

For each landing level $s=\sigma_{k+1}$, we take 256 training prompts with two independent noise pairs $(\eps,\bm\zeta)$ per prompt and sample $X$ from the teacher's default solver (28 Euler steps at $w=7$ on \sdthree, 50 at $w=4$ on FLUX, and 40 at $w=1$ on \qwen).

\begin{table}[t]
\centering
\footnotesize
\tabcolsep=2.7pt
\caption{\textbf{Calibration statistics and resulting query density.}
$d_s$ and $\hat b_s$ at $\tau=0.99$, in units of $\mathrm{err}_s$; mean query level of $\rho_k$ vs.\ uniform on $[s,0.99)$, and effective support on $[s,0.99)$.}
\label{tab:app-calib}
\begin{tabular}{@{}l r r r r r r@{}}
\toprule
 & & & & \multicolumn{2}{c}{Mean level} & \\
\cmidrule(lr){5-6}
Teacher & $s$ & $d_s(0.99)$ & $\hat b_s(0.99)$ & $\rho_k$ & uniform & Support \\
\midrule
\sdthree & 0.900 & 18.8 & 13.6 & 0.913 & 0.945 & 0.39 \\
 & 0.750 & 33.3 & 26.9 & 0.793 & 0.870 & 0.46 \\
 & 0.500 & 64.7 & 52.7 & 0.608 & 0.745 & 0.55 \\
\dashedmidrule{7}
\flux & 0.967 & 1.57 & 0.36 & 0.978 & 0.979 & 0.99 \\
 & 0.908 & 2.94 & 1.02 & 0.942 & 0.949 & 0.96 \\
 & 0.767 & 4.27 & 1.63 & 0.853 & 0.879 & 0.92 \\
\dashedmidrule{7}
\fluxnine & 0.967 & 1.52 & 0.33 & 0.978 & 0.979 & 0.99 \\
 & 0.908 & 2.84 & 0.87 & 0.942 & 0.949 & 0.96 \\
 & 0.767 & 4.37 & 1.53 & 0.853 & 0.879 & 0.92 \\
\dashedmidrule{7}
\qwen-2.1 & 0.857 & 4.64 & 1.94 & 0.915 & 0.923 & 0.95 \\
 & 0.666 & 7.66 & 3.10 & 0.811 & 0.828 & 0.94 \\
 & 0.400 & 15.0 & 5.73 & 0.663 & 0.695 & 0.92 \\
\bottomrule
\end{tabular}

\end{table}

On these coupled observations, define
\begin{equation}
\begin{aligned}
\Delta_{s,\tau}&=\Tw(\tilde\Y_\tau,\tau)-\Tw(\Y_s,s),\\
\mathrm{err}_s&=\E\|\Tc(\Y_s,s)-X\|^2,\\
d_s(\tau)&=\frac{\E\|\Delta_{s,\tau}\|^2}{\mathrm{err}_s},
\end{aligned}
\end{equation}
where norms are averaged over dimensions.
Unlike the raw discrepancy $d_s(\tau)$, which includes unavoidable averaging variance from added noise, we compute the projected score
\begin{equation}
\hat b_s(\tau)=\max_{\phi}\frac{\big(\E\langle\Delta_{s,\tau},\phi\rangle\big)^2}{\E\|\phi\|^2\,\mathrm{err}_s}
\end{equation}
across 13 levels in $(s,0.99]$.
The maximum ranges over nonzero features
\[
\phi\in\{\Tw,\Tc,\Tw-\Tc\}\quad\text{at }(\tilde\Y_\tau,\tau),
\]
reducing to $\Tc$ on \qwen where $w=1$.
Conditioning on the prompt $c$ as well, define the population conditional mean and bound the drift:
\[
\begin{aligned}
m_{s,\tau}&=\E[\Tw(\Y_s,s)\mid\tilde\Y_\tau,c],\\
\hat b_s(\tau)&\le\frac{\E\|\Tw(\tilde\Y_\tau,\tau)-m_{s,\tau}\|^2}{\mathrm{err}_s}.
\end{aligned}
\]
The conditional fluctuation around $m_{s,\tau}$ is orthogonal to every feature of $(\tilde\Y_\tau,c)$; Cauchy--Schwarz gives the bound.
We estimate these expectations from the calibration samples (\cref{sec:query}).
During training, we linearly interpolate the score and use inverse-CDF sampling from
\[
\rho_k(\tau)\propto\exp\!\big(-\hat b_{\sigma_{k+1}}(\tau)\big),\quad
\tau\in[\sigma_{k+1},0.99).
\]
\Cref{tab:app-calib} reports $d_s(0.99)$, $\hat b_s(0.99)$, the mean query level, and the normalized effective support $e^{H(\rho_k)}/(0.99-s)$ (with differential entropy $H$, so a uniform law has support 1): $\rho_k$ concentrates near $s$ on \sdthree and stays nearly uniform on FLUX and \qwen.

\section{Baselines}
\label{app:baselines}

\noindent\textbf{Official checkpoints.}
On \sdthree, we evaluate the official 4-step checkpoints of TDM~\cite{luo2025tdm} (with its recommended DPM-Solver++~\cite{lu2022dpmsolverpp} sampler) and CDM~\cite{liu2026cdm} (with its recommended Euler sampler) using their default inference settings at guidance $1$.
Similarly, the official 4-step FLUX.2-klein checkpoints are evaluated with their default pipelines at $w=1$.

\noindent\textbf{Local implementations and tuning.}
All locally trained baselines share our teacher-default guidance, batch size 16, rank-32 LoRA parameterization, and evaluation pipeline:
\begin{itemize}[leftmargin=*,itemsep=2pt,topsep=2pt]
\item \emph{DMD2} uses its distribution matching and backward simulation objectives (without the GAN loss) and five fake updates per student update; we lowered DMD2's default LoRA student learning rate to $2\times10^{-5}$ for stability with fake rate $5\times10^{-5}$ on \flux and \sdthree, and used student/fake rates of $10^{-6}/2.5\times10^{-5}$ on \fluxnine.
\item \emph{D-DMD} is reproduced from its paper and the implementation released with CDM~\cite{liu2026cdm}. Starting from CDM's default rates (student $10^{-5}$, fake $5\times10^{-6}$), we also tested a student rate of $5\times10^{-6}$, a fake rate of $2\times10^{-5}$, and a CFG-augmentation (CA) weight of $0.5$, retaining $5\times10^{-6}$ for both networks on \flux and \sdthree, and student/fake rates of $6.25\times10^{-7}/2.5\times10^{-6}$ on \fluxnine.
\item \emph{CDM} on \flux was tuned over LoRA student rates in $[10^{-6},10^{-5}]$ (constant and cosine), fake rates in $[5\times10^{-7},5\times10^{-5}]$, and CA weights in $\{0.5,1\}$, as well as full-parameter runs, which underperformed LoRA at batch size 16; \cref{tab:main} reports the best configuration (LoRA student $10^{-6}$, fake $5\times10^{-7}$, CA weight 0.5), and \fluxnine uses student/fake rates of $5\times10^{-7}/2.5\times10^{-7}$.
\item \emph{TDM} on \flux (student $4\times10^{-6}$, fake $2\times10^{-5}$) and \fluxnine (student $2\times10^{-6}$, fake $10^{-5}$) uses our $\ell_2$ re-implementation, which incorporates the DMD normalizer into an $\ell_2$ surrogate target, samples intervals in $\sigma$, and takes five fake updates per student update. This recipe outperforms TDM's official recipe (Huber loss divided by the normalizer and one fake update per student update; \cref{tab:app-reimpl}).
\end{itemize}

\begin{table}[!t]
\centering\footnotesize\tabcolsep=2.4pt
\caption{\textbf{Local reproductions.}
Peaks of local runs.
\Cref{tab:main} reports the official TDM and CDM checkpoints on \sdthree and the $\ell_2$ re-implementation of TDM (10.259) on \flux; App.~\ref{app:plugin} builds on the $\ell_2$ re-implementation.}
\label{tab:app-reimpl}
\resizebox{\columnwidth}{!}{%
\begin{tabular}{@{}l r r r r r r@{}}
\toprule
Implementation & HPSv3++ & HPSv3 & Pick & Aes. & HPSv2.1 & CLIP \\
\midrule
\multicolumn{7}{@{}l}{\emph{\sdthree, trained locally}} \\
\rowcolor{black!5}
TDM (official recipe) & 9.634 & 10.473 & 22.461 & 5.441 & 0.3002 & 0.3212 \\
\rowcolor{black!5}
TDM ($\ell_2$ re-impl.) & 9.993 & 10.510 & 22.397 & 5.527 & 0.3029 & 0.3181 \\
\rowcolor{black!5}
CDM & 10.586 & 11.787 & 22.554 & 5.469 & 0.3012 & 0.3210 \\
\midrule
\multicolumn{7}{@{}l}{\emph{\flux, trained locally}} \\
\rowcolor{black!5}
TDM (official recipe) & 9.333 & 9.098 & 22.224 & 5.446 & 0.2783 & 0.3145 \\
\bottomrule
\end{tabular}}

\end{table}

\noindent\textbf{Training prompts.}
All \sdthree and FLUX runs train on CDM's ${\sim}191\text{k}$-prompt corpus~\cite{liu2026cdm}; \qwen uses prompt-enhanced rewrites of a deduplicated ${\sim}190\text{k}$-prompt subset (App.~\ref{app:qwen}).
Overlap with the 200 DrawBench evaluation prompts is negligible (at most three prompts appear verbatim or as substrings).

\section{\qwen-2.1}
\label{app:qwen}

\begin{figure*}[t]
\centering
\includegraphics[width=\textwidth]{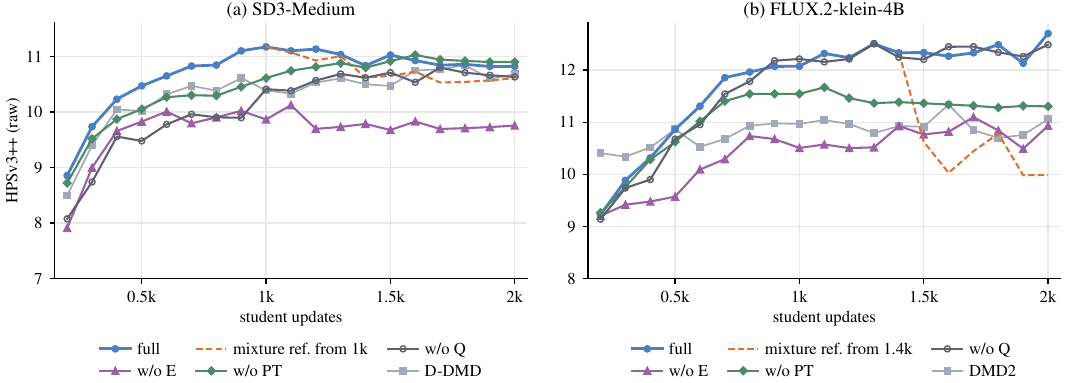}
\caption{\textbf{Training curves of the ablations.}
Raw \hpspp on DrawBench-1000 every 100 updates, with the strongest baseline of \cref{tab:main} (gray squares).
Dashed: continuation with the mixture reference from 1k (\sdthree) and 1.4k (\flux) updates.}
\label{fig:app-curves}
\end{figure*}

\noindent\textbf{Protocol and baselines.}
All \qwen runs (\cref{tab:qwen}) train and evaluate on H800 GPUs at the teacher's default $w=1$ using Qwen's official prompt-enhanced rewrites in the 1K resolution tier, with a 600-update budget and global batch size 16 (eight GPUs, two micro-batches).
During evaluation on DrawBench-1000, case $i$ is rendered with seed $42+i$ at the rewrite's aspect ratio and scored against the original DrawBench prompt (so scores are compared only within \cref{tab:qwen}).
Baselines are our \qwen adaptations of the released DMD2 and TDM recipes, evaluated under the same protocol.
TDM uses student and fake learning rates of $4\times10^{-6}$ and $2\times10^{-5}$, one fake update per student update, and a cosine schedule ending at 600 updates.
Turbo is the official 8-step Qwen-Image-2.1-Turbo pipeline.

\begin{table}[t]
\centering
\setlength{\tabcolsep}{2.2pt}
\caption{\textbf{\qwen-2.1 baselines with the PAG contrast of \ours.} Protocol and selection as in \cref{tab:qwen}, whose rows without PAG are repeated. \textbf{Bold}: best.}
\label{tab:app-qwen-pag}
\resizebox{\columnwidth}{!}{%
\begin{tabular}{@{}l c r r r r r r r@{}}
\toprule
Method & Ckpt & \shortstack{HPS\\v3++$\uparrow$} & \shortstack{HPS\\v3$\uparrow$} & Pick$\uparrow$ & Aes.$\uparrow$ & \shortstack{HPS\\v2.1$\uparrow$} & IR$\uparrow$ & CLIP$\uparrow$ \\
\midrule
DMD2 & 100R & 14.212 & 13.073 & 23.090 & 5.781 & 0.3032 & 1.049 & 0.3126 \\
DMD2 + PAG & 500E & 14.094 & 13.104 & 23.012 & \textbf{5.859} & 0.3061 & 1.047 & \textbf{0.3145} \\
TDM & 500R & 13.787 & 12.804 & 23.101 & 5.743 & 0.3003 & 1.072 & 0.3129 \\
TDM + PAG & 600R & 13.674 & \textbf{13.189} & \textbf{23.201} & 5.846 & 0.3003 & 1.048 & 0.3122 \\
\rowcolor{oursrow}
\ours & 400E & \textbf{14.691} & 13.186 & 23.175 & 5.788 & \textbf{0.3109} & \textbf{1.104} & 0.3138 \\
\bottomrule
\end{tabular}}

\end{table}

\noindent\textbf{\ours on \qwen.}
\ours follows \cref{tab:app-hyper} on \qwen's native 4-step shifted grid (which varies slightly with aspect ratio; the $1024^2$ landing levels are listed in \cref{tab:app-calib}), using a constant student learning rate after 100 warm-up updates and three adjustments.
First, the fake loss weight is $\min\{\mathrm{SNR}(\tau),5\}$.
Second, for $\tau\ge\sigma_{K-1}$, the teacher target uses the PAG contrast~\cite{ahn2024pag} of \cref{eq:contrast},
\begin{equation}
B=\Tc+\gamma\,\big(\Tc-\Tc^{\mathrm{pag}}\big),\qquad\gamma=5,
\end{equation}
where $\Tc^{\mathrm{pag}}$ perturbs self-attention in block 6.
For queries below $\sigma_{K-1}$, PT integrates $\Tc$.
Third, intermediate steps are queried by forward-noising their landing states $\z_{k+1}$ from $\sigma_{k+1}$ to $\tau$ via \cref{eq:forward}: with $a=(1-\tau)/(1-\sigma_{k+1})$,
\begin{equation}
\y=a\,\z_{k+1}+\sqrt{\tau^2-a^2\sigma_{k+1}^2}\;\bm\zeta,
\end{equation}
rather than re-noising $\xh_k$ with fresh noise (similarly to RTDMD~\cite{huang2026rtdmd}).
On \qwen, landing-state queries reduce visual artifacts observed under fresh-noise queries and improve peak \hpspp from 14.41 to 14.61 (in an ablation at $\gamma=3$).
On \sdthree and \flux, they lower peak \hpspp by 0.10 and 0.33, respectively, so we retain fresh-noise queries.
Calibration (App.~\ref{app:calib}) uses $w=1$ without PAG.

\noindent\textbf{Baselines with the same PAG.}
To separate the effect of PAG from the distillation objective, \cref{tab:app-qwen-pag} compares DMD2 and TDM with and without the PAG contrast used by \ours: block 6, $\gamma=5$ for $\tau\ge\sigma_{K-1}$.
Adding PAG improves some metrics but not peak \hpspp for either baseline.
\ours leads on \hpspp, HPSv2.1, and ImageReward.

\section{Training curves and run-to-run variability}
\label{app:curves}

\begin{figure*}[t]
\centering
\includegraphics[width=\textwidth]{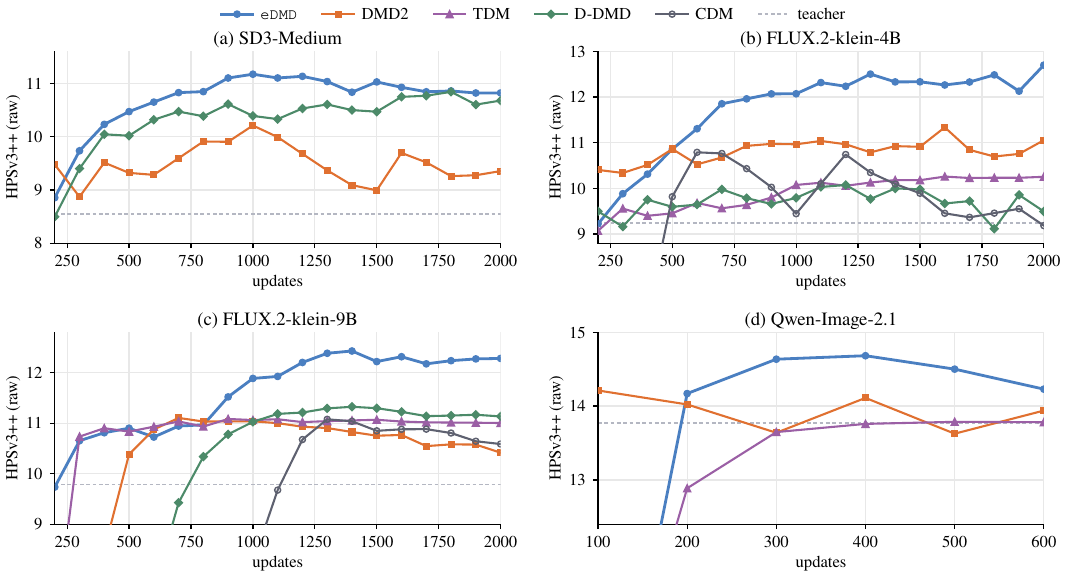}
\caption{\textbf{Training curves for the trained rows of \cref{tab:main,tab:qwen}.}
Raw weights, every 100 updates, unsmoothed; gray: the teacher's default sampler.
The official checkpoints (TDM and CDM on \sdthree, the FLUX.2-klein distilled models, \qwen-2.1-Turbo) have no training curve.
Values below the plotted range are clipped.}
\label{fig:app-allcurves}
\end{figure*}

\noindent\textbf{All methods.}
\Cref{fig:app-allcurves} shows raw \hpspp during training for every trained row of \cref{tab:main,tab:qwen}.
\Cref{tab:app-ckpt} lists each row's highest-scoring checkpoint within budget, selected over raw (R) and EMA (E) weights.

\begin{table}[t]
\centering\footnotesize\tabcolsep=6pt
\caption{\textbf{Checkpoints in \cref{tab:main,tab:qwen}.} R/E: raw/EMA; official: released checkpoint; `--': not run (App.~\ref{app:qwen}).}
\label{tab:app-ckpt}
\begin{tabular}{@{}l c c c c@{}}
\toprule
Method & SD3 & FLUX 4B & FLUX 9B & Qwen \\
\midrule
DMD2 & 1100E & 1600R & 700R & 100R \\
TDM & official & 1600R & 900R & 500R \\
D-DMD & 1800R & 1400E & 1400R & -- \\
CDM & official & 600R & 1300R & -- \\
\rowcolor{oursrow}
\ours & 1000R & 2000R & 1400R & 400E \\
\bottomrule
\end{tabular}

\end{table}

\noindent\textbf{Held-out selection.}
Because peak checkpoints in \cref{tab:main,tab:qwen} are selected and evaluated on the same 200 prompts, we assess selection bias across 1{,}000 random 100/100 prompt splits (keeping each prompt's five seeds together), selecting each trained row's checkpoint on one half and scoring it on the held-out half.
\ours ranks first on the held-out half in all 1{,}000 splits across all four teachers, leading the strongest baseline by a mean margin of 0.30, 1.37, 1.11, and 0.47 \hpspp on \sdthree, \flux, \fluxnine, and \qwen; across all rows, the largest split-averaged difference is 0.06.

\noindent\textbf{Ablation curves.}
\Cref{fig:app-curves} shows the raw \hpspp training curves of the ablations in \cref{tab:ablation}.
On \sdthree, \ours peaks at 1k updates; at 2k, its raw and EMA scores are 10.82 and 10.86, close to the best baseline peak (10.84).
On both FLUX models, its scores at 1k and 2k remain above every baseline's peak.

\noindent\textbf{The fake without E.}
The w/o E ablation replaces only the fake model's training data and loss with those of our TDM re-implementation (``TDM, released code'' in \cref{tab:app-tdmfake}).
For each sample, we draw a stage uniformly and re-noise its landing state via \cref{eq:app-kernel}.
We sample $\tau\sim\mathcal U[\max\{\sigma_{k+1},0.02\},\,\sigma_k-0.01)$ and weight the loss proportionally to $\min\{\mathrm{SNR}(\tau),5\}\,w^{1/n}$, where $w$ is the importance ratio in \cref{eq:app-tdmw}.
The regression target is $\xh_k$; at the final stage the landing state is $\x$ and $w=1$.
All student rollout, query, target, and update settings are retained.
On \fluxnine, this w/o E variant trails the full method by 0.53 on average from 300 updates onward (peak 11.54 vs.\ 12.43).
Replacing the endpoint fake with DMD2's unweighted mixture fake from scratch similarly degrades \flux performance, trailing by 1.59 on average from 300 onward (peak 10.79 vs.\ 12.70).

\noindent\textbf{Run-to-run variance.}
To quantify run-to-run variance in \cref{tab:ablation}, we trained the w/o Q variant twice from scratch: across matched raw and EMA checkpoints from 100 updates onward, score differences have standard deviations of 0.074 on \sdthree and 0.11 on \flux.
Across two additional seeds on \sdthree, \ours peaks at 10.98 and 11.01, giving a three-seed mean of $11.05\pm0.10$, above every baseline in \cref{tab:main}.

\noindent\textbf{Updating only the final step.}
To isolate the contribution of intermediate steps' local updates, we train with all student updates restricted to the final step ($\pi_{K-1}=1$), leaving intermediate steps to evolve solely through shared weights: \hpspp plateaus at 7.5--8.4 from 700 to 2k updates on \flux (vs.\ 12.1 at 1k and 12.7 at 2k for the full method) and peaks at 4.64 on \sdthree (vs.\ 11.17).

\begin{figure*}[p]
\centering
\includegraphics[width=\linewidth,height=0.86\textheight,keepaspectratio]{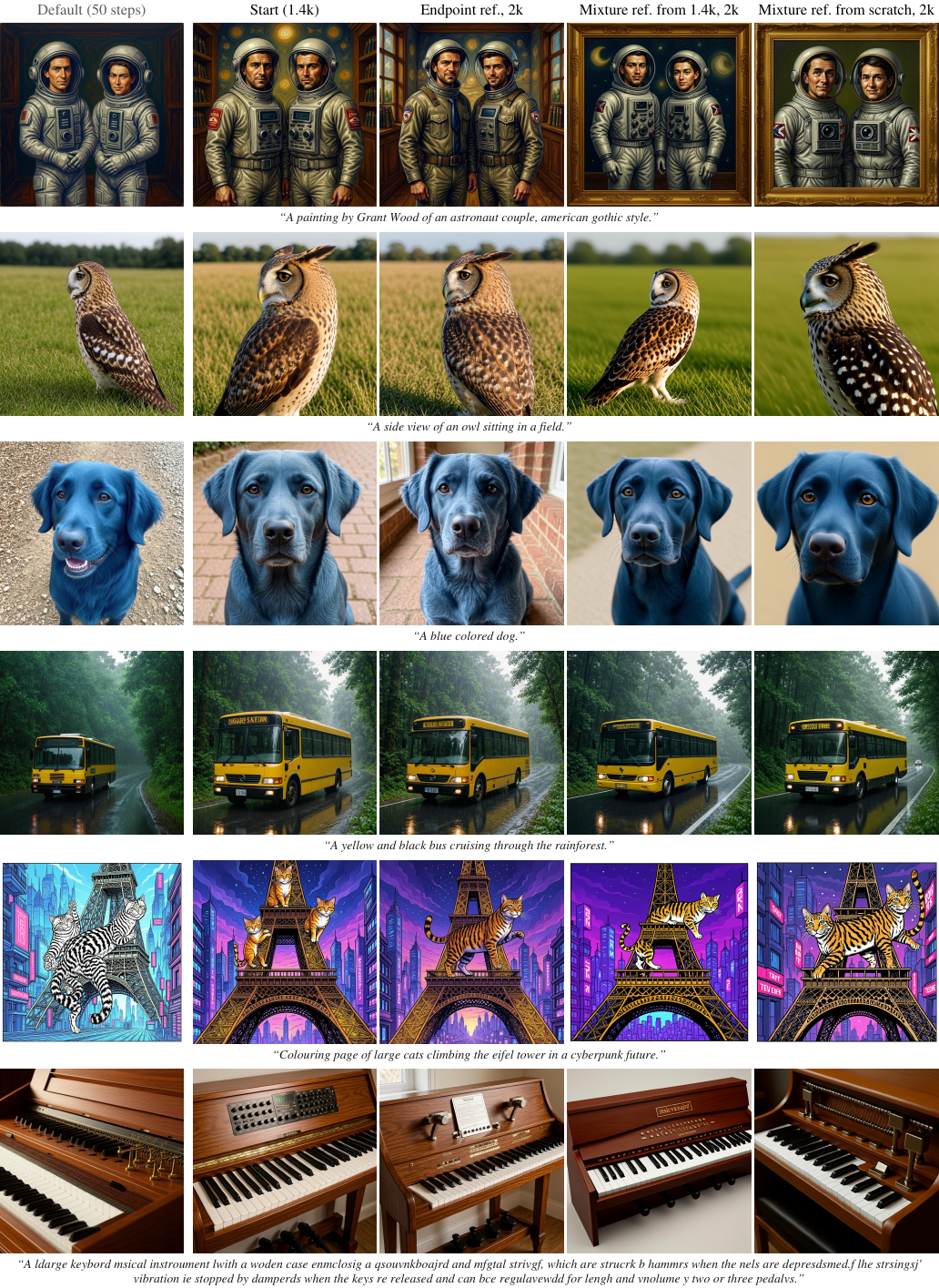}
\caption{\textbf{Switching a trained student to the mixture reference (\flux).}
Columns, from left to right: the teacher's default sampler (50 steps); the shared starting checkpoint at 1.4k updates; continuation with the endpoint reference to 2k; continuation with the mixture reference to 2k; and training with the mixture reference from scratch to 2k.
All students use four sampling steps and raw weights.
Each row shares a DrawBench prompt and initial noise (the bottom row is verbatim from DrawBench's Misspellings category).}
\label{fig:app-fork}
\end{figure*}

\noindent\textbf{The mixture reference on a trained student.}
We switch a trained \ours student to the mixture reference at 1k updates on \sdthree and 1.4k on \flux, keeping all other settings fixed.
On \flux, \hpspp drops by 1.71 within 100 updates and stays below the endpoint continuation through 2k; on \sdthree, the mean gaps are 0.22 raw and 0.12 EMA over 1.1k--2k.
\Cref{fig:app-fork} shows paired samples from these continuations: switching to the mixture reference produces smoother textures and simpler backgrounds in several examples, resembling the mixture-reference run trained from scratch.
To rule out a transient shock from an unadapted fake model, pre-fitting the \flux mixture fake for 1k updates with the student frozen, then resuming with identical inputs, still reduces raw \hpspp by 2.07 within 100 updates.

\noindent\textbf{Intermediate clean predictions.}
By \cref{prop:zerosets}, the mixture reference leaves a matched final output ($q=p$) unperturbed only when the intermediate predictions $\xh_0,\xh_1,\xh_2$ also mix to $p$.
As a simple diagnostic of whether real few-step students depart from matching the final output's scale, we measure the norm ratio $\|\xh_k\|/\|\xh_3\|$ over the first 64 DrawBench-1000 evaluation cases (raw weights).
Under the endpoint reference on \flux, the first two steps' mean ratios grow from $2.6$ and $2.2$ at 1k updates to $3.3$ and $2.6$ at 2k, showing that intermediate predictions need not match the scale of the final output.
By contrast, intermediate-based references pull these norms back toward $1$: at 2k updates on \flux, TDM keeps mean ratios at $1.15, 1.14, 0.99$, whereas adding the endpoint reference to TDM (\cref{app:plugin}) raises them to $2.51, 2.17, 1.24$; training from scratch under the mixture reference likewise keeps them within $1.3$.
Correspondingly, switching \ours to the mixture reference at 1.4k updates on \flux compresses the first two ratios from $2.93$ and $2.35$ to $1.43$ and $1.24$ within 100 updates alongside the 1.71 \hpspp drop, matching \cref{sec:ablation}.

\section{The endpoint reference on TDM}
\label{app:plugin}

\begin{table}[!t]
\centering
\setlength{\tabcolsep}{3pt}
\caption{\textbf{The endpoint reference added to TDM.}
\hpspp on DrawBench-1000.
E: endpoint reference; Peak: best within 2{,}000 updates over raw and EMA weights; @1k/@2k: raw weights.}
\label{tab:app-plugin}
\resizebox{\columnwidth}{!}{%
\begin{tabular}{@{}l ccc ccc ccc@{}}
\toprule
& \multicolumn{3}{c}{\sdthree} & \multicolumn{3}{c}{\flux} & \multicolumn{3}{c}{\fluxnine} \\
\cmidrule(lr){2-4}\cmidrule(lr){5-7}\cmidrule(l){8-10}
Run & Peak & @1k & @2k & Peak & @1k & @2k & Peak & @1k & @2k \\
\midrule
TDM re-impl. & 9.99 & 9.74 & 9.97 & 10.26 & 10.08 & 10.25 & 11.08 & 11.06 & 11.00 \\
TDM re-impl. + E & 10.60 & 10.46 & 10.59 & 11.70 & 11.50 & 11.58 & 11.62 & 11.27 & 11.59 \\
\midrule
\rowcolor{oursrow}
\ours & 11.17 & 11.17 & 10.82 & 12.70 & 12.07 & 12.70 & 12.43 & 11.89 & 12.28 \\
\bottomrule
\end{tabular}}

\end{table}

We replace the stage-wise fake of our TDM re-implementation (App.~\ref{app:baselines}) with the endpoint fake and train from scratch on \sdthree, \flux, and \fluxnine (\cref{tab:app-plugin}).
The fake model is trained solely on final outputs $\x$ noised to $\tau\sim\mathcal U[0.02,0.99)$ with TDM's clean-prediction target and loss weight $\min\{\mathrm{SNR}(\tau),5\}$ (so TDM's re-noised-state importance weight drops out), while retaining TDM's Euler rollout, stage-interval queries, normalizer, student loss, optimizers, and five fake updates per student update.
The endpoint reference raises peak \hpspp by 0.61 on \sdthree, 1.44 on \flux, and 0.54 on \fluxnine, with gains persisting at both 1k and 2k updates.

\section{Qualitative results}
\label{app:qual}

\begin{figure*}[p]
\centering
\includegraphics[width=\linewidth,height=0.86\textheight,keepaspectratio]{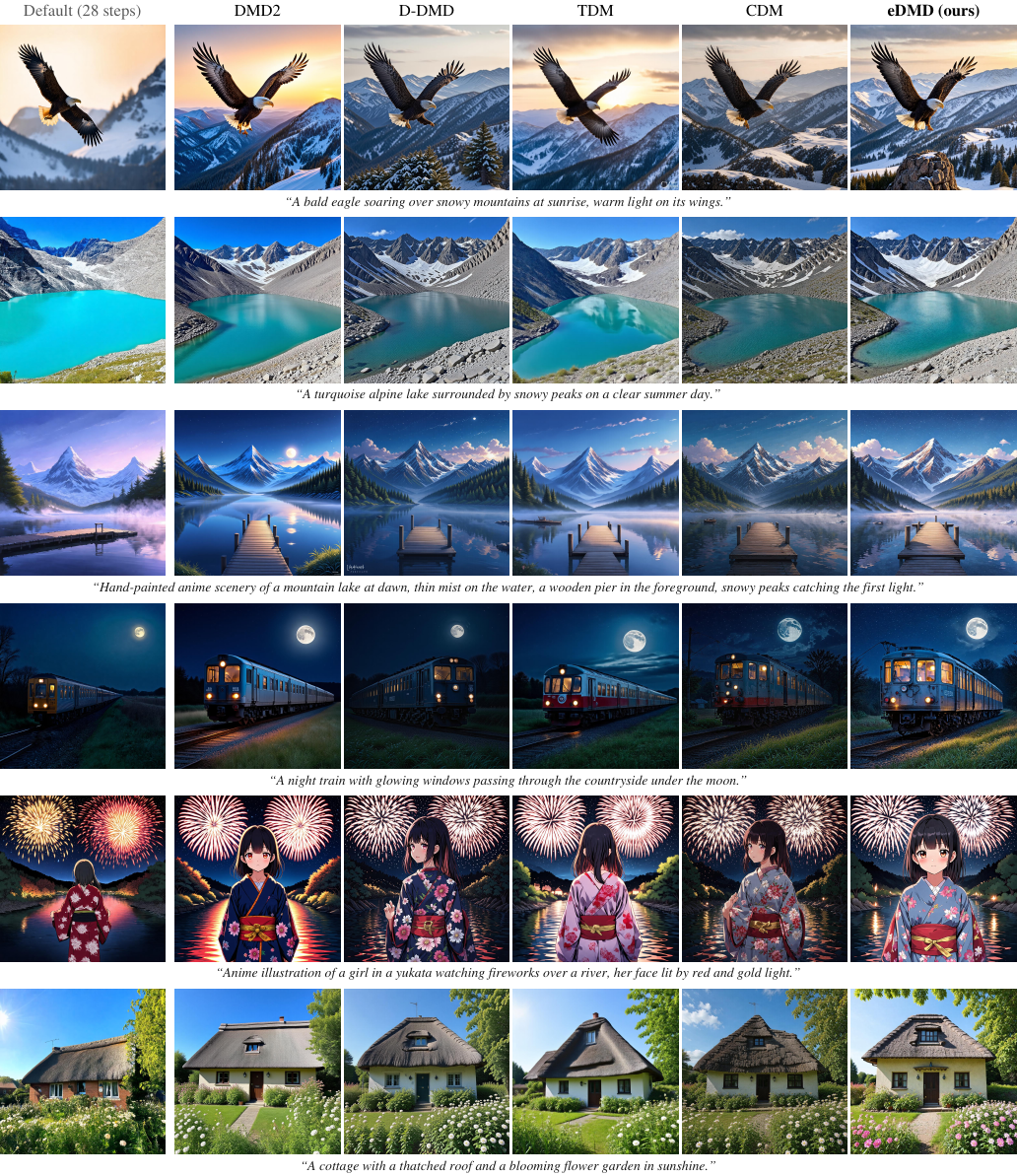}
\caption{\textbf{Qualitative results on \sdthree.} Leftmost: teacher's default sampler (28 steps, $w=7$); other columns: 4-step students of \cref{tab:main}.}
\label{fig:qual-sd3-cmp}
\end{figure*}

\begin{figure*}[p]
\centering
\includegraphics[width=\linewidth,height=0.86\textheight,keepaspectratio]{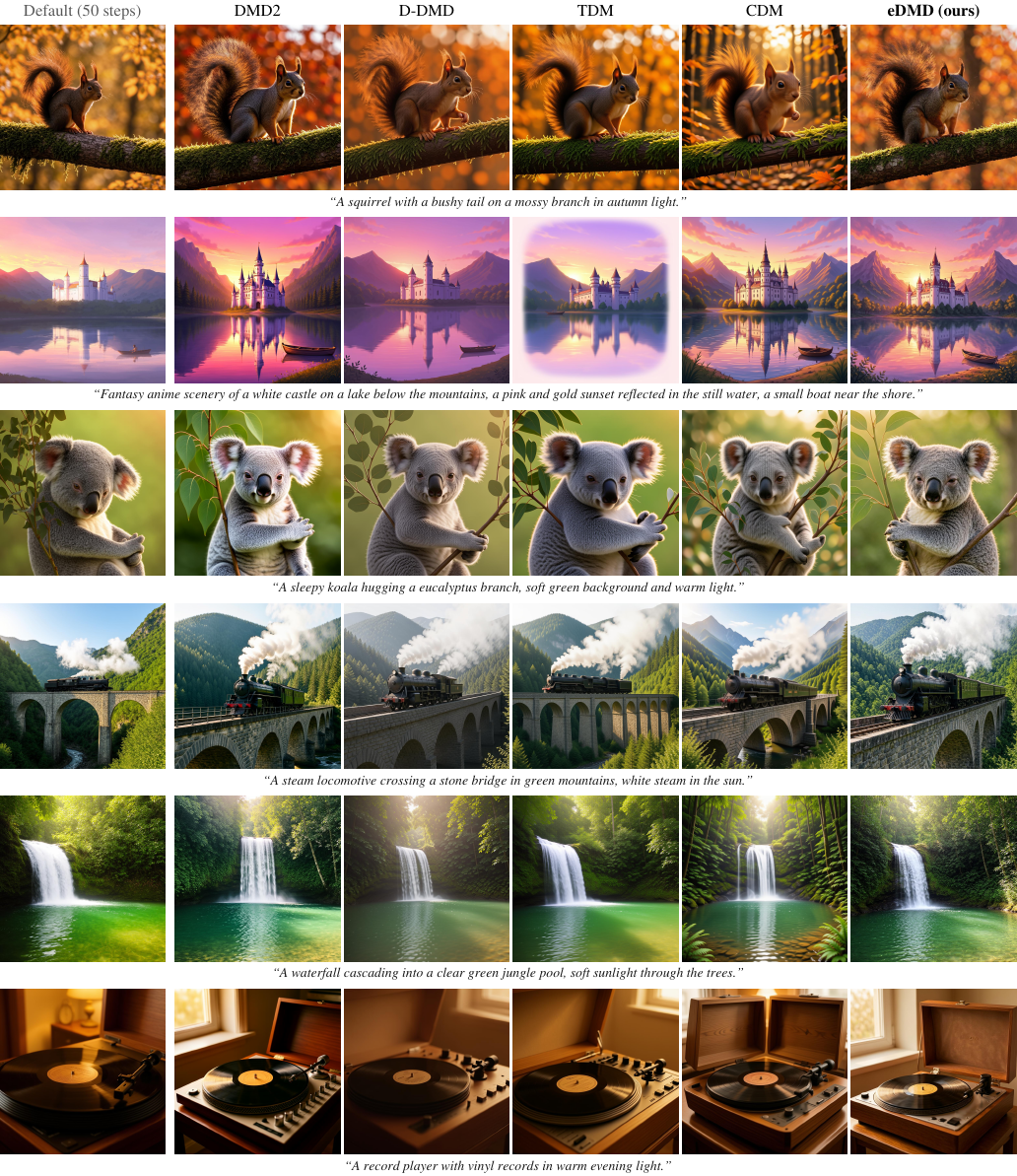}
\caption{\textbf{Qualitative results on \flux.} Leftmost: teacher's default sampler (50 steps, $w=4$); other columns: 4-step students of \cref{tab:main}.}
\label{fig:qual-flux-cmp}
\end{figure*}

\begin{figure*}[p]
\centering
\includegraphics[width=\linewidth,height=0.86\textheight,keepaspectratio]{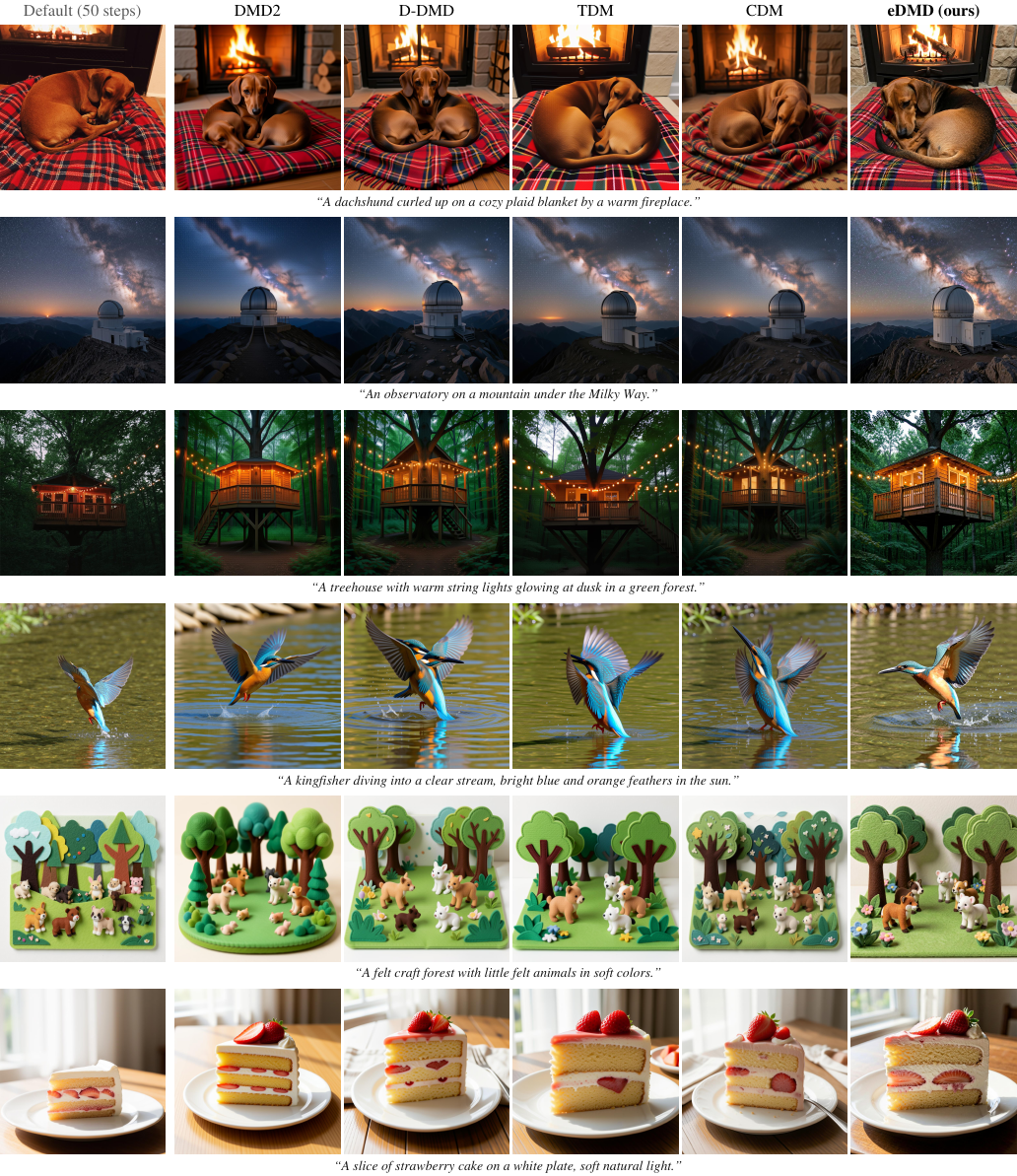}
\caption{\textbf{Qualitative results on \fluxnine.} Leftmost: teacher's default sampler (50 steps, $w=4$); other columns: 4-step students of \cref{tab:main}.}
\label{fig:qual-flux9b-cmp}
\end{figure*}

\begin{figure*}[p]
\centering
\includegraphics[width=\linewidth,height=0.86\textheight,keepaspectratio]{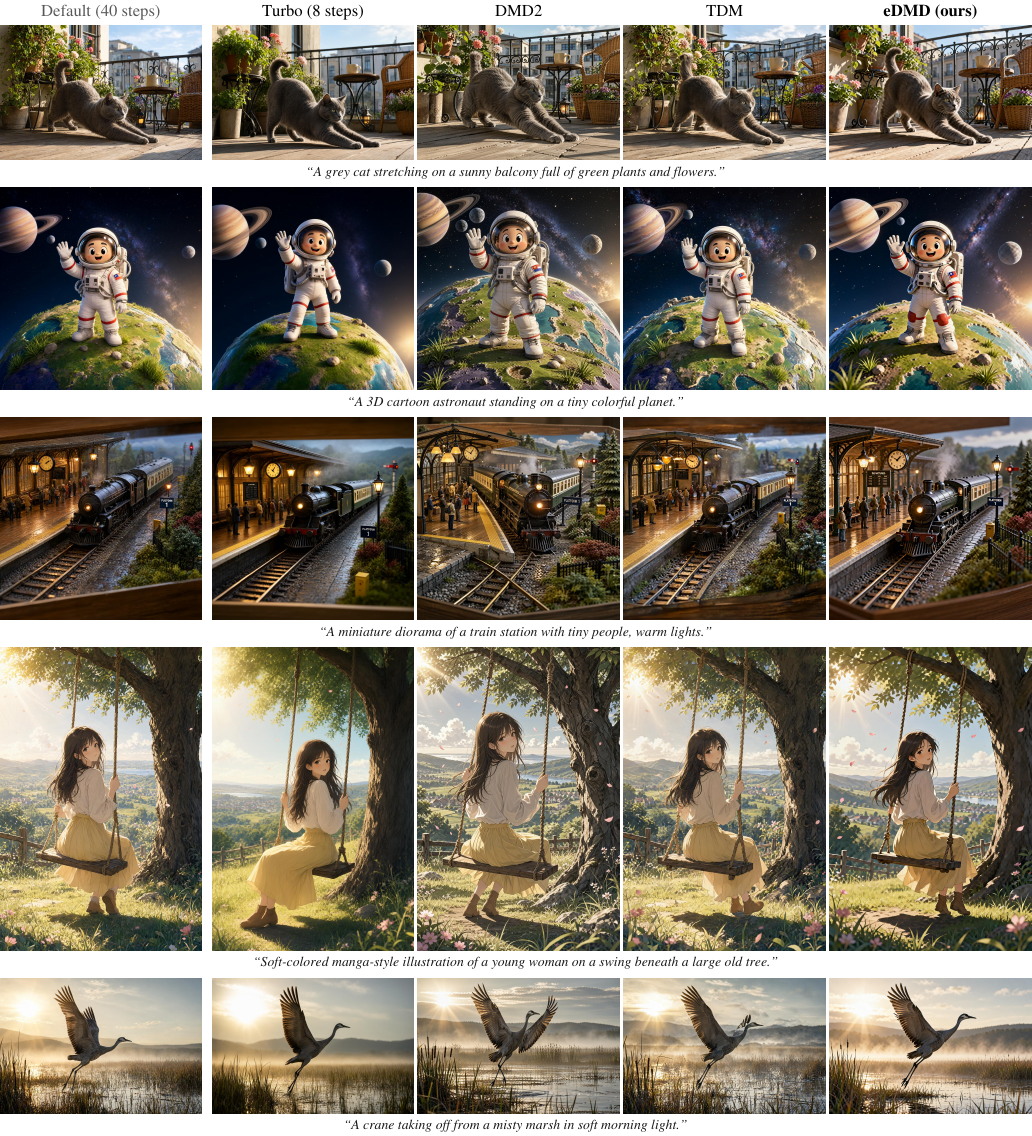}
\caption{\textbf{Qualitative results on \qwen-2.1.} Leftmost: teacher's default sampler (40 steps); Turbo: official 8-step distillation; other columns: 4-step students, all rendered from the same prompt-enhanced rewrite at its aspect ratio with the same seed.}
\label{fig:qual-qwen}
\end{figure*}

\Cref{fig:qual-sd3-cmp,fig:qual-flux-cmp,fig:qual-flux9b-cmp,fig:qual-qwen} show more comparisons on each teacher.
Each row shares a prompt and initial noise.
The leftmost column uses the teacher's default sampler; the others use student checkpoints from \cref{tab:main} (\qwen: \cref{tab:qwen}).

\end{document}